\documentclass[10pt,twocolumn,letterpaper]{article}

\usepackage[pagenumbers]{cvpr} 

\usepackage{subcaption}
\usepackage{amsfonts}
\usepackage{amsmath}
\usepackage{booktabs}
\usepackage{makecell}
\usepackage{capt-of}
\usepackage{algorithm}
\usepackage{algorithmic}

\definecolor{cvprblue}{rgb}{0.21,0.49,0.74}
\usepackage[pagebackref,breaklinks,colorlinks,allcolors=cvprblue]{hyperref}

\def\paperID{*****} 
\def\confName{CVPR}
\def\confYear{2026}

\title{PoseGen: In-Context LoRA Finetuning for Pose-Controllable Long Human Video Generation}

\author{
Jingxuan He$^{1,2, \ast}$ \quad
Busheng Su$^{1}$ \quad
Finn Wong$^{1, \dagger}$\\
$^{1}$Xiaoice, China\\
$^{2}$The University of Sydney, Australia\\
{\tt\small jihe0215@uni.sydney.edu.au, bushengsu@gmail.com, fantasyfw@gmail.com}
}

\begin{document}

\twocolumn[{%
\renewcommand\twocolumn[1][]{#1}%
\maketitle
\begin{center}
    \includegraphics[width=1.0\textwidth]{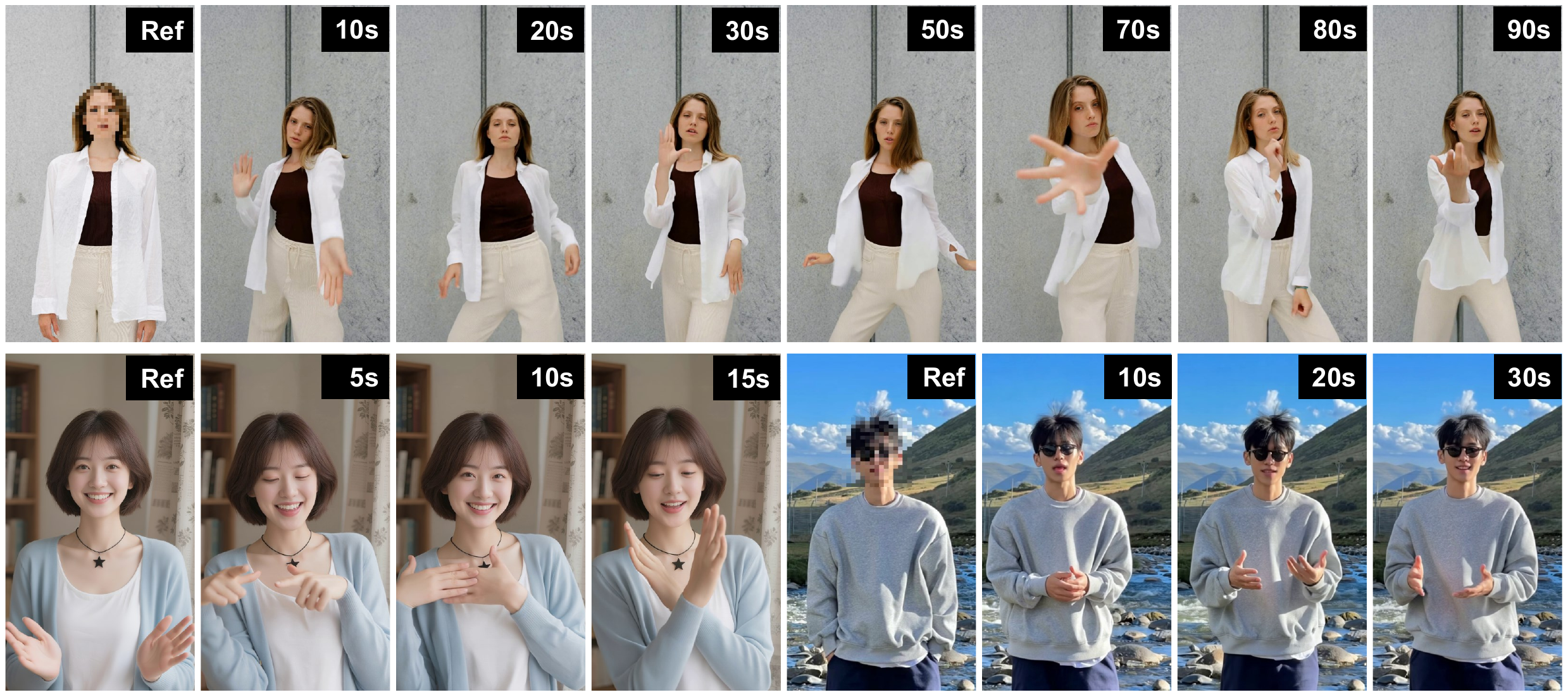}
    \captionof{figure}{We present \textbf{PoseGen}, a pose-controllable human video generation framework capable of generating long videos from a human reference image and a driving pose sequence. Our framework maintains visual fidelity to the reference appearance while ensuring temporal coherence throughout the whole synthesized video.}
    \label{fig:teaser}
\end{center}%
}]

\let\thefootnote\relax\footnotetext{$^{\ast}$Work primarily done while at Xiaoice.}
\let\thefootnote\relax\footnotetext{$^{\dagger}$Corresponding author.}

\begin{abstract}
Generating temporally coherent, long-duration videos with precise control over subject identity and movement remains a fundamental challenge for contemporary diffusion-based models, which often suffer from identity drift and are limited to short video length. We present PoseGen, a novel framework that generates human videos of extended duration from a single reference image and a driving video. Our contributions include an in-context LoRA finetuning design that injects subject appearance at the token level for identity preservation, while simultaneously conditioning on pose information at the channel level for fine-grained motion control. To overcome duration limits, we introduce a segment-interleaved generation strategy, where non-overlapping segments are first generated with improved background consistency through a shared KV-cache mechanism, and then stitched into a continuous sequence via pose-aware interpolated generation. Despite being trained on a remarkably small 33-hour video dataset, PoseGen demonstrates superior performance over state-of-the-art baselines in identity fidelity, pose accuracy, and temporal consistency. Code is available at \url{https://github.com/Jessie459/PoseGen}.
\end{abstract}
\section{Introduction}

The synthesis of high-fidelity video content has emerged as a key frontier in generative AI, propelled by the remarkable success of diffusion models in image generation ~\cite{rombach2022high}.
The ability to automatically create realistic and controllable videos holds transformative potential across a wide range of applications, including entertainment, filmmaking, virtual reality, and digital marketing.
In particular, there is a critical demand for methods that generate videos with faithful identity preservation and precise motion control.

This demand has motivated extensive research into controllable video generation~\cite{li2024dispose,zhang2025mimicmotion,zhou2024realisdance,tu2025stableanimator,gan2025humandit,hu2025hunyuancustom,jiang2025vace}.
Despite the progress, challenges persist in controllability and duration.
Existing methods typically suffer from three limitations:
1) Identity Drift: the subject's appearance morphs over time;
2) Motion Inaccuracy: precise motion control without visual artifacts remains difficult to achieve;
3) Limited Duration: most models are architecturally confined to short clips (typically under 10 seconds), with longer generation leading to severe accumulation errors.
Moreover, a significant yet overlooked challenge is the substantial compute cost and data appetite of state-of-the-art approaches.
Many leading models require large-scale proprietary datasets that exceed 10K hours of video.
This high resource barrier limits accessibility for the broader research community and hinders practical deployment.

To address these challenges, we propose PoseGen, a novel framework for pose-controllable, long-duration video generation from a single reference image.
PoseGen builds upon a pretrained video diffusion model~\cite{wan2025wan} and leverages the efficiency of LoRA~\cite{hu2022lora} for lightweight adaptation.
As part of our contributions, we introduce a dual conditioning mechanism, termed in-context LoRA finetuning, to facilitate both identity preservation and motion control.
For identity preservation, we encode the reference image into VAE latents and concatenate them with the noise latents along the token dimension, thereby injecting explicit appearance context into the generation process.
For motion control, we concatenate pose skeleton and hand normal latents with the noise latents along the channel dimension, enabling precise frame-by-frame guidance and improving motion fidelity.

The primary contribution of our framework lies in its ability to generate long-duration videos with minimal identity drift and strong temporal consistency.
This is achieved via a segment-interleaved generation strategy.
Specifically, we first generate multiple non-overlapping short segments, where background consistency among them is maintained by caching and reusing the Key-Value (KV) pairs from the self-attention layers of a source segment~\cite{cao2023masactrl,cai2025ditctrl}.
These segments are then stitched together following the corresponding driving signal to enable seamless transitions across time.
To realize this strategy, we employ two LoRA modules with distinct roles.
The first is optimized for generating non-overlapping segments, while the second focuses on stitching adjacent segments to produce a temporally coherent and complete video.

Our contributions can be summarized as follows:
\begin{itemize}
    \item We propose PoseGen, an effective and efficient framework for generating long-duration human videos from a reference image and a driving video.
    \item We introduce a dual in-context conditioning mechanism that injects appearance identity at the token level and pose guidance at the channel level, achieving superior subject fidelity and motion control.
    \item We present a segment-interleaved generation strategy that leverages KV-sharing for improved background consistency, which effectively enhances temporal coherence for long video generation.
    \item We demonstrate through comprehensive quantitative and qualitative experiments that PoseGen outperforms state-of-the-art pose-driven human video generation approaches in terms of visual quality, identity fidelity, and long-duration temporal consistency.
\end{itemize}

\begin{figure*}[t]
    \centering
    \begin{subfigure}[t]{0.68\textwidth}
        \includegraphics[height=6.6cm]{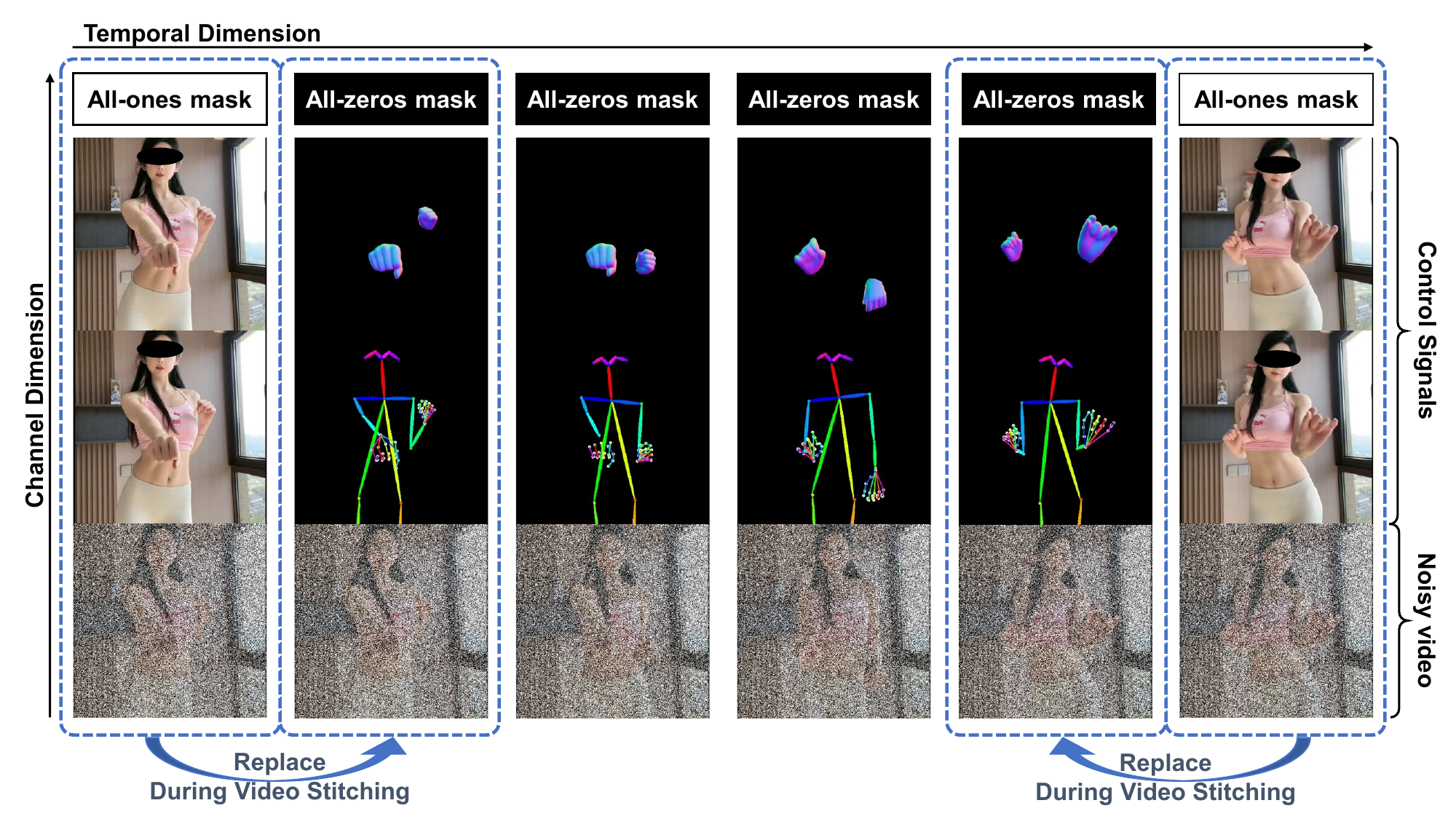}
        \caption{}
        \label{fig:framework}
    \end{subfigure}
    \hfill
    \begin{subfigure}[t]{0.31\textwidth}
        \includegraphics[height=6.6cm]{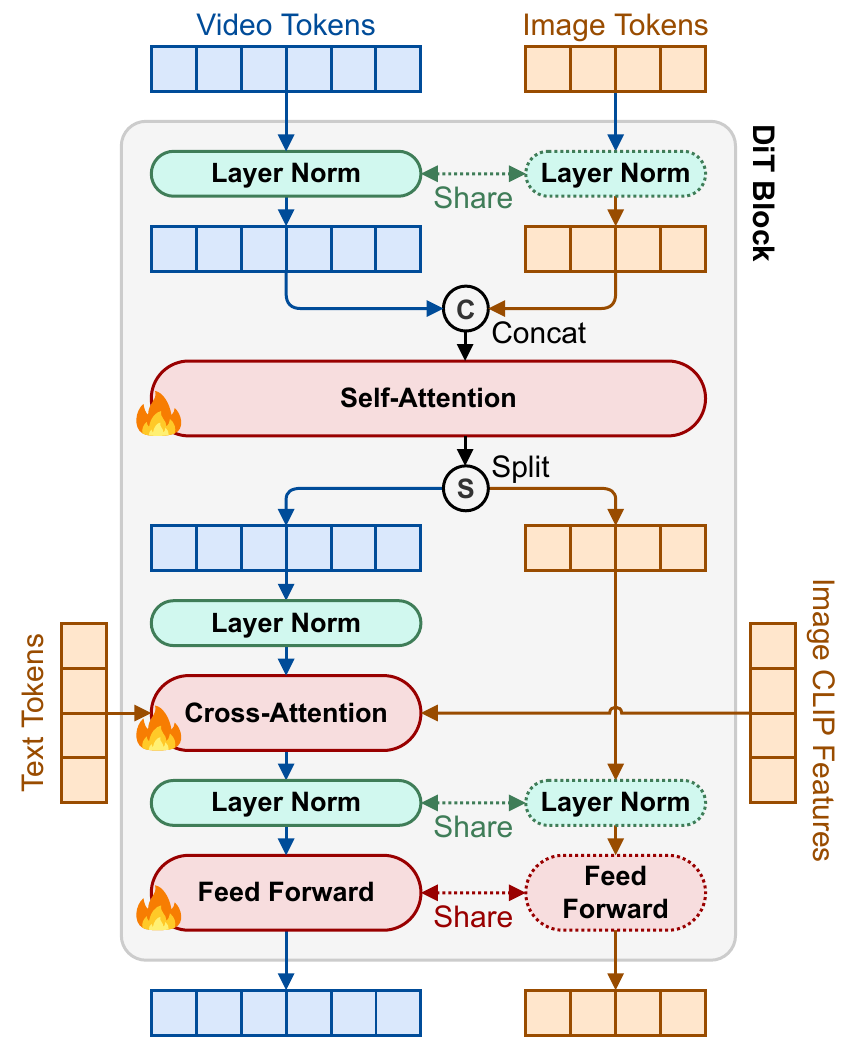}
        \caption{}
        \label{fig:block}
    \end{subfigure}
    \caption{Overview of in-context LoRA finetuning. (a) Motion control. We adopt pose skeletons and hand surface normals as control signals. These conditions are concatenated with noisy video frames in the latent space. (b) Reference injection. Image and video tokens are processed through a stack of DiT blocks with shared parameters. LoRA is applied to the self-attention, cross-attention, and feed forward layers in each DiT block.}
    \label{fig:lora_finetunning}
\end{figure*}
\section{Related Work}

\noindent \textbf{Identity-consistent video generation.}
Maintaining subject identity from a reference image is a core challenge in video synthesis.
Early techniques like DreamBooth~\cite{ruiz2023dreambooth} finetunes the entire model for a specific subject, resulting in high storage and slow adaptation.
Recent parameter-efficient approaches such as DreamActor-M1~\cite{luo2025dreamactor} and VACE~\cite{jiang2025vace} inject identity features via tailored attention mechanisms.
OmniHuman-1~\cite{lin2025omnihuman} and HunyuanVideo-Avatar~\cite{hu2025HunyuanVideo-Avatar} build audio-driven human-centric systems using omni-condition mixed training and face-aware masking adapters, respectively.
However, these methods often rely on intricate architectural designs or complex training schemes.
In contrast, PoseGen employs in-context LoRA finetuning to efficiently preserve identity in long videos.

\noindent \textbf{Pose-controllable video generation.}
Controlling subject movement is essential for creating dynamic and purposeful content.
Inspired by ControlNet~\cite{zhang2023adding}, a significant line of work conditions video diffusion models on explicit control signals such as skeletal poses or edge maps.
AnimateDiff~\cite{guo2023animatediff} introduce a series of motion modules that can be seamlessly plugged into existing text-to-image models to animate images.
Animate Anyone~\cite{hu2024animate} and UniAnimate~\cite{wang2024unianimate} design separate encoders for appearance and pose injection.
Recent models like X-Dyna~\cite{chang2025x} introduce dynamics adapters to balance background context with fluid effects, while others explore DiTs~\cite{zhou2025realisdance,gan2025humandit,peebles2022DiT} for improved scalability.
Despite these advances, the high computational cost of modeling long temporal sequences remains a major bottleneck.
In contrast, PoseGen discards heavyweight pose encoders and employs a channel-wise conditioning mechanism for precise motion control.
We also design an interleaved generation strategy to achieve long-range motion control without modifying the architecture.

\noindent \textbf{Long video generation.}
Generating videos beyond a few seconds remains challenging due to computational and memory constraints.
Recent efforts explore autoregressive models~\cite{yin2025slow}, which show promising results but suffer from inherent error accumulation.
To mitigate this, methods such as MAGI-1~\cite{teng2025magi} and SkyReels-V2~\cite{chen2025skyreels} introduce controlled noise during training and perform autoregressive denoising for long video generation.
However, these methods often require training from scratch on large-scale datasets, limiting their adaptability to downstream tasks.
Training-free methods~\cite{qiu2023freenoise,kim2024fifo} manipulate injected noise but still remain prone to error accumulation and exhibit degraded visual quality and temporal coherence.
In pose-controllable human video generation, most existing methods~\cite{hu2024animate,zhang2025mimicmotion,wang2025unianimate} extend video length by synthesizing overlapping segments and merging them, which often leads to boundary inconsistencies.
Our method addresses this issue through a segment-interleaved generation mechanism that produces non-overlapping segments with consistent backgrounds and then stitches them via pose-aware interpolation.
\section{Method}

Given a reference image and a driving video, PoseGen synthesizes high-fidelity videos that faithfully replicate the driving motion while maintaining rigorous human identity and background consistency.

\subsection{Preliminaries}
Our method is built on Diffusion Transformers (DiTs)~\cite{peebles2023scalable}, which combine strong generation capabilities of diffusion models~\cite{sohl2015deep, ho2020denoising} with the expressiveness and scalability of Transformers~\cite{vaswani2017attention}. 
Modern diffusion models designed for video synthesis~\cite{yang2024cogvideox, wan2025wan} typically employ a causal 3D Variational Auto-Encoder (VAE) to compress videos from pixel space to latent space, in order to reduce computational complexity and memory consumption. 
The 3D VAE consists of an encoder $\mathcal{E}$ and a decoder $\mathcal{D}$, where the encoder compresses a video $\mathbf{x} \in \mathbb{R}^{3 \times F \times H \times W}$ into its corresponding video latents $\mathbf{z}=\mathcal{E}(\mathbf{x}) \in \mathbb{R}^{c \times f \times h \times w}$ and the decoder projects the video latents back to the original pixels $\mathbf{x}=\mathcal{D}(\mathbf{z}) \in \mathbb{R}^{3 \times F \times H \times W}$. 
Here $F$, $H$, and $W$ denote the number of frames, height, and width, respectively.
For Wan2.1~\cite{wan2025wan}, the latents have $c=16$ channels, with a temporal length of $f = (F - 1)/4 + 1$, latent height $h = H/8$, and latent width $w = W/8$.

A DiT-based denoising model comprises three major modules: a patchifier, a stack of Transformer blocks, and an unpatchifier. 
The patchifier applies 3D convolutions with a kernel size of $(1, 2, 2)$ to embed video latents $\mathbf{z}$ into a sequence of video tokens $\mathbf{t} \in \mathbb{R}^{n \times d}$, where $d$ is the token embedding dimension, and $n=(f/1) \cdot (h/2) \cdot (w/2)$. 
Additionally, text prompts are embedded through cross-attention following the self-attention operation in each block.

\subsection{In-Context LoRA Finetuning}
In-context LoRA finetuning is a dual conditioning mechanism that simultaneously facilitates pose-driven human animation and visual appearance preservation.
It is a simple yet effective solution with minimal architectural modifications to the underlying diffusion model.

\noindent \textbf{Motion control.}
To achieve pose-driven motion control, we adopt pose skeleton images as the primary driving signals following established practices in prior works~\cite{hu2024animate, zhang2025mimicmotion, tu2025stableanimator}. 
However, in our early experiments, we observed that the synthesized videos often suffered from degraded quality in hand regions due to their high-frequency textures and rapid movements. 
To mitigate this issue, as presented in Fig.~\ref{fig:lora_finetunning} (a), we introduce hand surface normals as auxiliary control signals.
These normals are estimated using a surface normal prediction model in combination with a body-part segmentation model~\cite{khirodkar2024sapiens}.
This approach is advantageous as surface normals provide rich geometric cues, effectively resolving complex scenarios like overlapping hands where mesh-based estimation typically fails~\cite{zhou2024realisdance}.
To condition the diffusion model on these control signals, we concatenate pose skeleton and hand normal images with the noisy video along the channel dimension in latent space. 
Additionally, to enable pose-aware frame interpolation for long video generation, we introduce a masking mechanism to specify which frames are to be synthesized. 
Specifically, we adopt a binary mask $\mathbf{m} \in \{0, 1\}^{s \times f \times h \times w}$ in which a value of $0$ indicates frames to be generated and a value of $1$ signifies frames to be preserved.
Here $s$ is the temporal stride of the causal 3D VAE.
In the first phase where non-overlapping segments are generated, the mask $\mathbf{m}$ is filled with zeros; in the second phase where segments are interpolated, the positions of the first and last few frames are marked with ones.
Given the noisy video latents $\mathbf{z}_{\text{vid}} \in \mathbb{R}^{c \times f \times h \times w}$, pose skeleton latents $\mathbf{z}_{\text{pose}} \in \mathbb{R}^{c \times f \times h \times w}$, and hand normal latents $\mathbf{z}_{\text{hand}} \in \mathbb{R}^{c \times f \times h \times w}$, the video tokens $\mathbf{t}_{\text{vid}} \in \mathbb{R}^{n_{\text{vid}} \times d}$ with sequence length $n_{\text{vid}} = (f/1) \cdot (h/2) \cdot (w/2)$ are obtained as follows:
\begin{equation}
    \mathbf{t}_{\text{vid}} = \operatorname{Patchifier} \left( [\mathbf{z}_{\text{vid}}; \mathbf{m}; \mathbf{z}_{\text{pose}}; \mathbf{z}_{\text{hand}}] \right),
\end{equation}
where $[;]$ denotes concatenation along the channel axis, and $\operatorname{Patchifier}()$ is a patch embedding layer instantiated with 3D convolutions.

\noindent \textbf{Reference injection.}
To ensure that the synthesized video remains consistent with the subject's appearance depicted in the reference image, we propose an architecture-efficient method to inject the reference information into the modeling process. 
We first compress the reference image into its image latents $\mathbf{z}_{\text{img}} \in \mathbb{R}^{c \times 1 \times h \times w}$ using the same pretrained VAE. 
A separate patchifier is applied to convert the image latents $\mathbf{z}_{\text{img}}$ into image tokens $\mathbf{t}_{\text{img}} \in \mathbb{R}^{n_{\text{img}} \times d}$ with sequence length $n_{\text{img}} = (h/2) \cdot (w/2)$.
As shown in Fig.~\ref{fig:lora_finetunning} (b), image tokens $\mathbf{t}_{\text{img}}$ and video tokens $\mathbf{t}_{\text{vid}}$ are processed in parallel through a stack of DiT blocks. 
In each DiT block, two sets of tokens pass through self-attention and feed forward layers that share the same parameters. 
During self-attention, we concatenate video and image tokens along the sequence dimension and perform full self-attention. 
Specifically, video and image tokens are linearly projected to queries $\mathbf{Q}$, keys $\mathbf{K}$, and values $\mathbf{V}$.
Rotary Position Embedding (RoPE)~\cite{su2024roformer} is subsequently applied to queries and keys to provide positional information. 
Inspired by OminiControl~\cite{tan2024ominicontrol}, we shift the position indices of image tokens to avoid overlap with those of the video tokens, considering that the reference image and the synthesized frames are not strictly spatially aligned. 
The full self-attention of layer $l$ is then computed as:
\begin{equation}\label{eq:attention}
    \operatorname{Attn}([\mathbf{t}_{\text{vid}}^{(l)}; \mathbf{t}_{\text{img}}^{(l)}]) = \operatorname{Softmax} \left( \frac{ \mathbf{Q}^{(l)} {\mathbf{K}^{(l)}}^{\top} }{ \sqrt{d} } \right) \mathbf{V}^{(l)}.
\end{equation}
To enhance semantic understanding of the visual content, the video tokens additionally interact with text embeddings and CLIP~\cite{radford2021learning} features of the reference image after full self-attention.

\noindent \textbf{LoRA finetuning.}
Our approaches to motion control and reference injection inherently support parameter-efficient LoRA~\cite{hu2022lora} finetuning.
First, the channel-wise concatenation used in motion control imposes spatial alignment between guidance and generation, which effectively reduces the complexity of the learning process. 
Second, the sequence-wise concatenation adopted in reference injection enables flexible token-level interactions, leveraging the strong contextual modeling capabilities of the DiT architecture.
During finetuning, we optimize LoRA parameters applied to self-attention, cross-attention, and feed forward layers in DiT blocks, as well as the newly introduced parameters in patchifiers.

\begin{algorithm}[t]
    \small
    \caption{Inference-Time Background KV-sharing}
    \label{alg:kv_sharing}
    \textbf{Input}: Initial video tokens $\mathbf{t}_{\text{vid}}$; cached source keys $\{\mathbf{K}_{\text{src}}^{(l,t)}\}$ and source values $\{\mathbf{V}_{\text{src}}^{(l,t)}\}$
    \begin{algorithmic}[1]
        \FOR{$t=T, T-1, \dots, 1$}
            \FOR{$l=1, 2 \dots, L$}
                \STATE $\hat{\mathbf{t}}^{(l,t)}_{\text{vid}} \leftarrow \operatorname{Attn}([\mathbf{t}_{\text{vid}}^{(l,t)}; \mathbf{t}_{\text{img}}^{(l,t)}])$ \COMMENT{Eq.~(\ref{eq:attention})}
                \IF{Background KV-Sharing}
                    \STATE $\mathbf{M}^{(l,t)} \leftarrow \mathbf{Q}^{(l,t)}_{\text{text}}, \mathbf{K}^{(l,t)}_{\text{vid}}$ \COMMENT{Eq.~(\ref{eq:attention_mask})}
                    \STATE $\mathbf{M}^{(l,t)}_{\text{src}} \leftarrow \mathbf{Q}^{(l,t)}_{\text{text,src}}, \mathbf{K}^{(l,t)}_{\text{vid,src}}$ \COMMENT{Eq.~(\ref{eq:attention_mask})}
                    \STATE $\hat{\mathbf{t}}^{(l,t)}_{\text{vid,src}} \leftarrow \operatorname{Attn}([\mathbf{t}^{(l,t)}_{\text{vid}}; \mathbf{t}_{\text{img}}^{(l,t)}]; \mathbf{M}^{(l,t)}_{\text{src}} )$ \COMMENT{Eq.~(\ref{eq:masked_attention})}
                    \STATE Update $\hat{\mathbf{t}}^{(l,t)}_{\text{vid}}$ \COMMENT{Eq.~(\ref{eq:attention_fusion})}
                \ENDIF
                \STATE $\mathbf{t}_{\text{vid}}^{(l+1,t)} \leftarrow \operatorname{FeedForward} \left( \hat{\mathbf{t}}^{(l,t)}_{\text{vid}} \right)$
            \ENDFOR
            \STATE $\mathbf{t}_{\text{vid}}^{(1,t-1)} \leftarrow \mathbf{t}_{\text{vid}}^{(L,t)}$
        \ENDFOR
    \end{algorithmic}
    \textbf{Return}: Denoised video tokens $\mathbf{t}_{\text{vid}}$
\end{algorithm}

\subsection{Long Video Generation}

A key challenge in generating long videos is maintaining continuity across different segments.
Naively composing a long video from sequentially generated short clips is insufficient for two main reasons.
First, pose skeletons alone lack appearance details to ensure smooth transitions, often resulting in motion artifacts.
Second, independently generated segments suffer from background drift as the subject moves, disrupting the naturalness of a continues scene.

To address these issues, we introduce a novel segment-interleaved generation strategy that leverages background KV-sharing.
The process begins by first generating disjoint video segments at regular intervals. Then, we generate the ``in-between'' segments using a specialized LoRA model trained for interpolation, effectively stitching the base segments together into a single, seamless video.

The core idea for maintaining background consistency is to cache and reuse background key-value pairs from the self-attention layers of a source segment (e.g., the first generated segment). 
To identify background regions, we derive binary masks from attention maps computed using queries and keys in cross-attention layers.
Mathematically, given text queries $\mathbf{Q}_{\text{text}}^{(l)} \in \mathbb{R}^{n_{\text{text}} \times d}$, video keys $\mathbf{K}_{\text{vid}}^{(l)} \in \mathbb{R}^{n_{\text{vid}} \times d}$, and subject-relevant (e.g., ``woman'', ``man'') text token indices $\mathbb{I} \subset \{0, \dots, n_{\text{text}}-1\}$, the attention map $\mathbf{A}^{(l)}$ and the corresponding attention mask $\mathbf{M}^{(l)}$ are computed as:
\begin{align}
    \mathbf{A}^{(l)} &= \frac{1}{|\mathbb{I}|} \sum_{i \in \mathbb{I}} \left( \frac{ \mathbf{Q}_{\text{text}, i}^{(l)} {\mathbf{K}_{\text{vid}}^{(l)}}^{\top} }{ \sqrt{d} } \right), \label{eq:attention_map} \\
    \mathbf{M}^{(l)} &= \frac{1}{l} \sum_{i=1}^{l} \operatorname{Thresholding} \left( \mathbf{A}^{(i)} \right). \label{eq:attention_mask}
\end{align}
The resulting mask $\mathbf{M}^{(l)}$ takes the value of 1 for subject regions and 0 for the background via Otsu's thresholding. 
Inspired by mutual self-attention~\cite{cao2023masactrl}, we reformulate the existing self-attention in Eq.~(\ref{eq:attention}) to facilitate information exchange between current and source segments. 
Specifically, in the $l$-th self-attention layer, the queries $\mathbf{Q}^{(l)}$ remain unchanged, while the keys and the values are substituted with those from the corresponding self-attention layer of the source segment (denoted as $\mathbf{K}^{(l)}_{\text{src}}$ and $\mathbf{V}^{(l)}_{\text{src}}$):
\begin{equation}\label{eq:masked_attention}
    \begin{split}
        & \operatorname{Attn}([\mathbf{t}^{(l)}_{\text{vid}}; \mathbf{t}_{\text{img}}^{(l)}]; \mathbf{M}^{(l)}_{\text{src}} ) =\\
        & \left[ \left(1 - \mathbf{M}^{(l)}_{\text{src}}\right) \odot \operatorname{Softmax} \left( \frac{ \mathbf{Q}^{(l)} {\mathbf{K}^{(l)}_{\text{src}}}^{\top} }{ \sqrt{d} } \right) \right] \mathbf{V}^{(l)}_{\text{src}},
    \end{split}
\end{equation}
where $\mathbf{M}^{(l)}_{\text{src}}$ denotes the attention mask from the source segment computed analogously to $\mathbf{M}^{(l)}$ in Eq.~(\ref{eq:attention_mask}). 
$\mathbf{M}^{(l)}_{\text{src}}$ is introduced to exclusively extract background visual content. 
Subsequently, let $\hat{\mathbf{t}}^{(l)}_{\text{vid}}$ represent the updated video token embeddings after the self-attention layer, the embeddings are fused as follows:
\begin{equation}\label{eq:attention_fusion}
    \begin{split}
        \hat{\mathbf{t}}^{(l)}_{\text{vid}} \leftarrow &\hat{\mathbf{t}}^{(l)}_{\text{vid}} \odot \left(\mathbf{M}^{(l)} \lor \mathbf{M}_{\text{src}}^{(l)} \right) +\\
        &\hat{\mathbf{t}}^{(l)}_{\text{vid,src}} \odot \left(1 - \mathbf{M}^{(l)} \lor \mathbf{M}_{\text{src}}^{(l)} \right),
    \end{split}
\end{equation}
where $\odot$ is the element-wise product, $\lor$ is the OR operation to identify the common background in two segments.
The background KV-sharing method is summarized in Alg.~\ref{alg:kv_sharing}.

\section{Experiments}

\begin{figure*}[tb]
    \centering
    \includegraphics[width=0.95\linewidth]{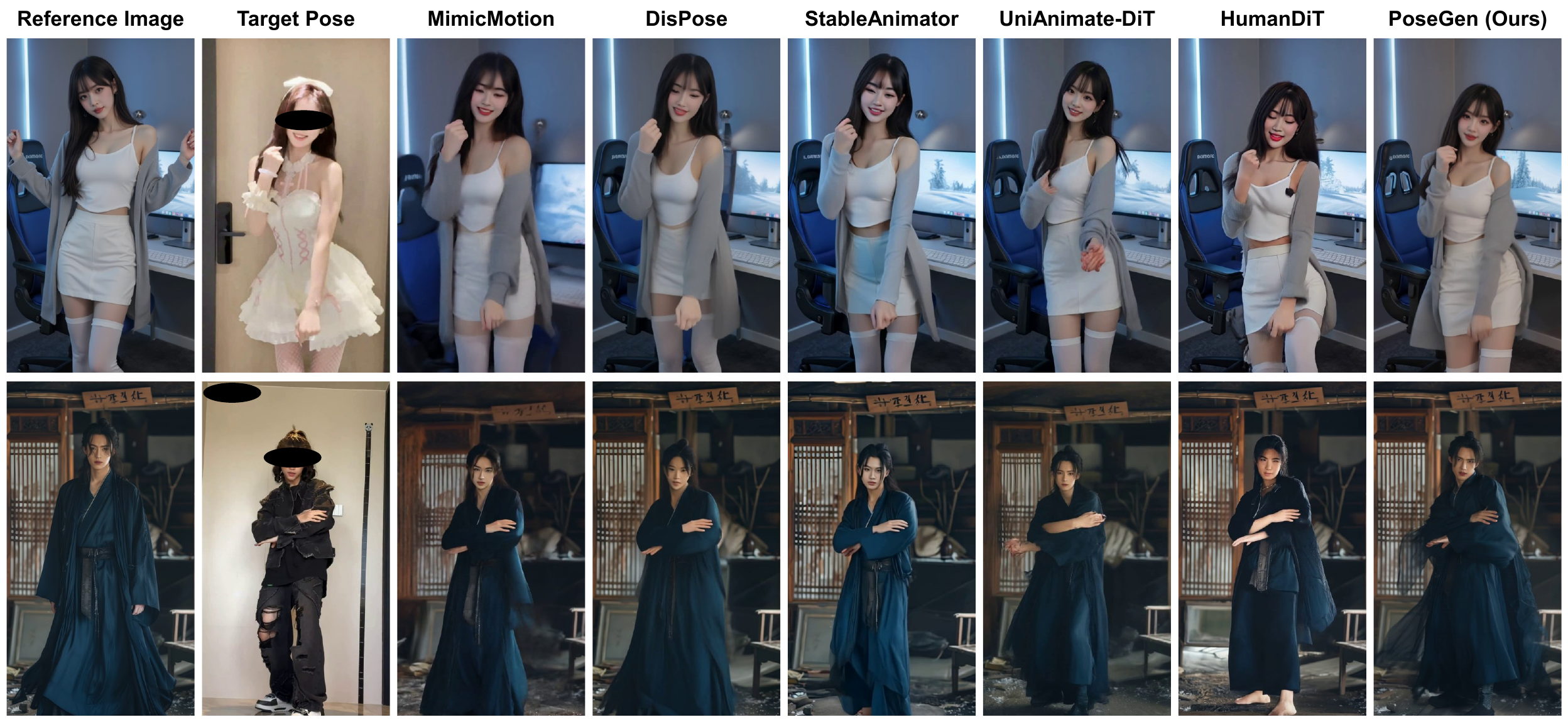}
    \caption{Qualitative comparison with state-of-the-art methods. The examples are from PoseGenVal (CrossID). Our method demonstrates superior performance in preserving the subject's identity and detailed appearance.}
    \label{fig:qualitative_results}
\end{figure*}

\subsection{Experimental Settings}

\noindent \textbf{Implementation details.}
We adopt Sapiens~\cite{khirodkar2024sapiens} for both pose skeleton and surface normal extraction.
The video generation model is based on the open-source backbone Wan2.1-I2V-14B~\cite{wan2025wan}.
We finetune all linear layers within self-attention, cross-attention, and feed forward modules using LoRA~\cite{hu2022lora} with a rank of 16 on a self-collected 33-hour video dataset (7,005 videos), which is orders of magnitude smaller than previous works~\cite{luo2025dreamactor, gan2025humandit}.
The first-phase LoRA was trained for 4,000 steps on $8$ NVIDIA A100 GPUs with a batch size of $8$ at $1280 \times 720$ spatial resolution.
The second-phase LoRA was trained to retain the first and last 1/4 frames while keeping other settings unchanged.
Background KV-sharing is performed at initial denoising steps and deep layers following the configuration in DiTCtrl~\cite{cai2025ditctrl}.
For details of our training data pre-processing, please refer to the Appendix.

\noindent \textbf{Evaluation metrics.}
We employ a comprehensive suite of image-level metrics (L1, PSNR, SSIM, LPIPS, FID) and video-level metrics (FID-VID, FVD)~\cite{hore2010image, wang2004image, zhang2018unreasonable, heusel2017gans, balaji2019conditional, unterthiner2018towards} to evaluate video generation quality.
For perceptual studies without ground truths, we measure visual quality via Q-Align~\cite{wu2023q}, identity preservation via ArcFace~\cite{deng2019arcface} feature similarity, and temporal consistency via DINO~\cite{caron2021emerging} feature similarity.

\subsection{Comparison with State-of-the-art Methods}

\begin{table*}[t]
    \centering
    \small
    \setlength{\tabcolsep}{5pt}
    \caption{Quantitative comparison results with state-of-the-arts. Each entry shows evaluation scores on TikTok (left) and PoseGenVal (SameID) (right). For all metrics, $\uparrow$ indicates higher is better, and $\downarrow$ indicates lower is better.}
    \begin{tabular}{l|ccccccc}
        \toprule
        Method & L1 (1E-5) $\downarrow$ & PSNR $\uparrow$ & SSIM $\uparrow$ & LPIPS $\downarrow$ & FID $\downarrow$ & FID-VID $\downarrow$ & FVD $\downarrow$ \\
        \midrule
        MimicMotion~\cite{zhang2025mimicmotion}    & 3.20/3.33 & 17.79/17.85 & 0.775/0.760 & 0.251/0.292 & 47.36/63.05 & 20.58/20.00 & 348.26/381.67 \\
        DisPose~\cite{li2024dispose}               & 3.09/3.31 & 18.04/18.07 & 0.794/0.783 & 0.247/0.276 & 50.91/67.88 & 16.83/24.08 & 387.37/409.85 \\
        StableAnimator~\cite{tu2025stableanimator} & 3.42/4.06 & 17.26/16.55 & 0.755/0.719 & 0.264/0.310 & 51.74/67.33 & 18.89/22.98 & 460.35/436.79 \\
        UniAnimate-DiT~\cite{wang2025unianimate}   & 2.03/2.15 & 20.59/17.54 & \textbf{0.828}/0.794 & 0.194/0.286 & 43.19/\textbf{52.36} & 29.40/19.98 & 280.59/282.43 \\
        HumanDiT~\cite{gan2025humandit}            & 2.17/\textbf{1.61} & \textbf{21.79}/19.78 & 0.824/\textbf{0.821} & \textbf{0.152}/\textbf{0.241} & 38.97/62.48 & 10.62/15.77 & 216.20/349.05 \\
        \textbf{PoseGen (Ours)} & \textbf{1.96}/\textbf{1.61} & 20.91/\textbf{20.24} & 0.814/\textbf{0.821}  & 0.172/0.242 & \textbf{38.67}/53.63 & \textbf{10.25}/\textbf{14.21} & \textbf{210.75}/\textbf{214.51} \\
        \bottomrule
    \end{tabular}
    \label{tab:same_identity_results}
\end{table*}

\begin{table}[t]
    \centering
    \small
    \setlength{\tabcolsep}{3pt}
    \caption{Quantitative comparison results with state-of-the-arts on PoseGenVal (CrossID). Each entry shows model-predicted (left) and human-evaluated (right) scores.}
    \begin{tabular}{l|cccccc}
        \toprule
         Method & \makecell{Visual \\ Quality} $\uparrow$ & \makecell{Identity \\ Preservation} $\uparrow$ & \makecell{Temporal \\ Consistency} $\uparrow$ \\
        \midrule
        MimicMotion       & 78.49/2.52 & 84.02/2.43 & 93.33/2.76 \\
        DisPose           & 79.48/2.71 & 87.01/2.49 & 93.71/2.89 \\
        StableAnimator    & 86.38/2.87 & 84.81/2.54 & 92.33/2.73 \\
        UniAnimate-DiT    & 87.22/3.55 & 94.75/3.36 & 94.85/3.08 \\
        HumanDiT          & 86.31/3.45 & 84.70/2.79 & 94.85/3.53 \\
        \textbf{PoseGen (Ours)} & \textbf{87.47/4.16} & \textbf{94.99/4.07} & \textbf{95.22/4.16} \\     
        \bottomrule
    \end{tabular}
    \label{tab:cross_identity_results}
\end{table}

We compare PoseGen with state-of-the-art pose-driven human video generation methods spanning both UNet-based and DiT-based architectures, including MimicMotion~\cite{zhang2025mimicmotion}, DisPose~\cite{li2024dispose}, StableAnimator~\cite{tu2025stableanimator}, UniAnimate-DiT~\cite{wang2025unianimate}, and HumanDiT~\cite{gan2025humandit}.
Experiments are conducted on the TikTok dataset~\cite{jafarian2021learning} following previous works~\cite{zhang2025mimicmotion, li2024dispose, tu2025stableanimator}.
To further evaluate model performance beyond short clips in TikTok, we construct two validation datasets, PoseGenVal (SameID) and PoseGenVal (CrossID), where each video contains 400-600 frames.
PoseGenVal (SameID) includes nine human movement videos and is introduced to assess long-term temporal coherence with available ground-truths.
PoseGenVal (CrossID) serves as a more challenging set that comprises thirteen image-video pairs, where each reference image depicts a distinct identity sourced from a mixture of real and AI-generated photos.
Notably, our introduced validation datasets contain a comparable number of samples to TikTok, but feature longer videos and greater diversity across close-, medium-, and long-shot compositions.

\noindent \textbf{Quantitative results.}
\cref{tab:same_identity_results} shows quantitative results on TikTok and PoseGenVal (SameID).
As illustrated, PoseGen demonstrates superior performance over current state-of-the-art methods across most metrics.
Specifically, it not only achieves the lowest L1 error, indicating superior pixel-wise accuracy, but also significantly outperforms competitors on video-level metrics (FID-VID and FVD) by notable margins.
This highlights its strong capability of maintaining temporal consistency in perceptually realistic videos.
Evaluations on the more challenging PoseGenVal (CrossID) (see \cref{tab:cross_identity_results} left-side entries) further substantiate PoseGen's strengths.
In these cross-identity scenarios, our framework consistently surpasses all baselines across three evaluated perceptual dimensions.
For instance, while UniAnimate-DiT remains competitive in terms of visual quality and identity preservation, PoseGen's pronounced advantage in temporal consistency directly validates the effectiveness of our proposed long-video generation strategy.

\noindent \textbf{Qualitative results.}
As shown in Fig.~\ref{fig:qualitative_results}, PoseGen excels at preserving subject identity and appearance details where other state-of-the-art methods exhibit visual degradation.
For instance, competing methods may struggle to accurately replicate clothing texture and color (top row), or exhibit distortions in complex traditional attire and facial features (bottom row). In contrast, PoseGen faithfully preserves these intricate details while ensuring coherent motion, yielding visually superior and more authentic results.

\noindent \textbf{User study.}
To evaluate perceptual quality, we conducted a user study comparing PoseGen against five other leading competitors on PoseGenVal (CrossID). 
For this study, we invited 27 external participants to rate generated cross-identity videos on a scale of 0-5 in terms of three criteria: visual quality, identity preservation, and temporal consistency. Results summarized in \cref{tab:cross_identity_results} (right-side entries) reveal a clear preference for PoseGen. Our method achieves the highest scores across all categories by a large margin, demonstrating its superior ability to generate high-quality, coherent, and perceptually appealing videos.

\subsection{Ablation Study}

\begin{figure}[tb]
    \centering
    \includegraphics[width=1.0\linewidth]{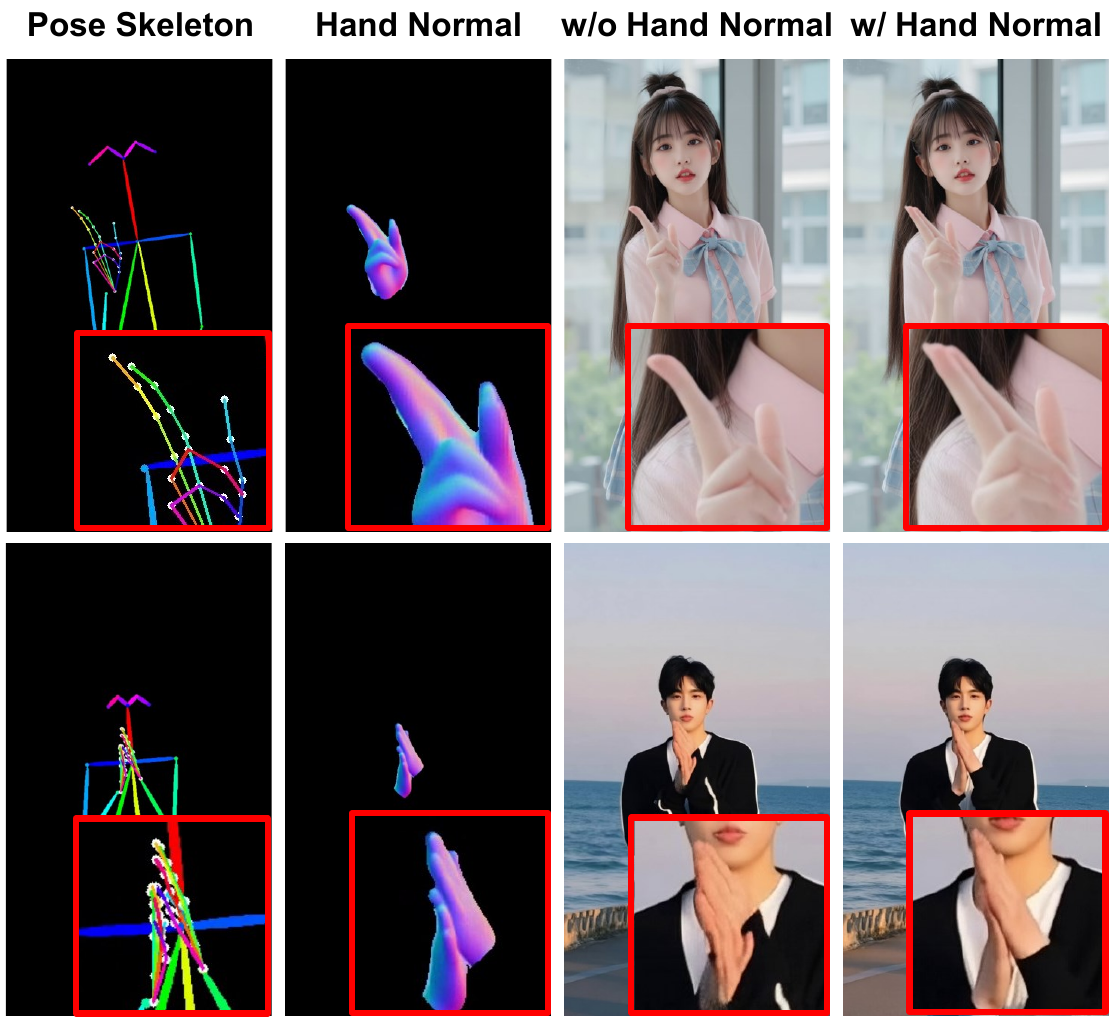}
    \caption{Visualization of the effectiveness of incorporating hand surface normals. Red boxes enlarge hand regions of interested.}
    \label{fig:ablation_hand}
\end{figure}

\begin{table}[tb]
    \centering
    \small
    \caption{Ablation on pose adherence in terms of Percentage of Correct Keypoints (PCK) with the threshold 5\%.}
    \begin{tabular}{lcc}
        \toprule
        Pose Skeleton                            & Hand Normal  & PCK@0.05${\uparrow}$ \\
        \midrule
        $\checkmark$ (DWPose~\cite{yang2023effective})          & $\times$     & 0.779 \\
        $\checkmark$ (Sapiens-Pose~\cite{khirodkar2024sapiens}) & $\times$     & 0.791 \\
        $\checkmark$ (Sapiens-Pose~\cite{khirodkar2024sapiens}) & $\checkmark$ & \textbf{0.809} \\
        \bottomrule
    \end{tabular}
    \label{tab:ablation_pose_adherence}
\end{table}

\noindent \textbf{Pose adherence.}
We initially followed prior works~\cite{zhang2025mimicmotion,li2024dispose,tu2025stableanimator} and adopted DWPose~\cite{yang2023effective} as the sole pose-driving condition.
However, we noticed that DWPose was occasionally inaccurate and therefore replaced it with the more robust Sapiens-Pose~\cite{khirodkar2024sapiens} for keypoint prediction.
Additionally, we observed that perceptual artifacts predominantly occurred in hand regions. 
To mitigate this issue, we propose to incorporate hand surface normals as an additional driving signal beyond pose skeletons.
To evaluate the accuracy of pose adherence, we perform keypoint estimation on both real and generated videos from the PoseGenVal dataset and calculate Percentage of Correct Keypoints (PCK) between them.
The results shown in \cref{tab:ablation_pose_adherence} demonstrate that employing Sapiens-Pose with hand normals yields the best performance.
Although the improvement in terms of PCK introduced by hand surface normals is relatively small, we hypothesize that this is due to the limited spatial extent of hand regions.
We further present illustrative comparisons in \cref{fig:ablation_hand}.
The example in the first row shows that the dense surface normals complement the sparse pose skeletons by providing nuanced shape information, which effectively prevents the unintended merging of distinct fingers.
The example in the second row demonstrates that surface normals help disambiguate interactions between hands by capturing fine-grained geometric details, resulting in anatomically plausible hand positions.

\begin{figure}[tb]
    \centering
    \includegraphics[width=1.0\linewidth]{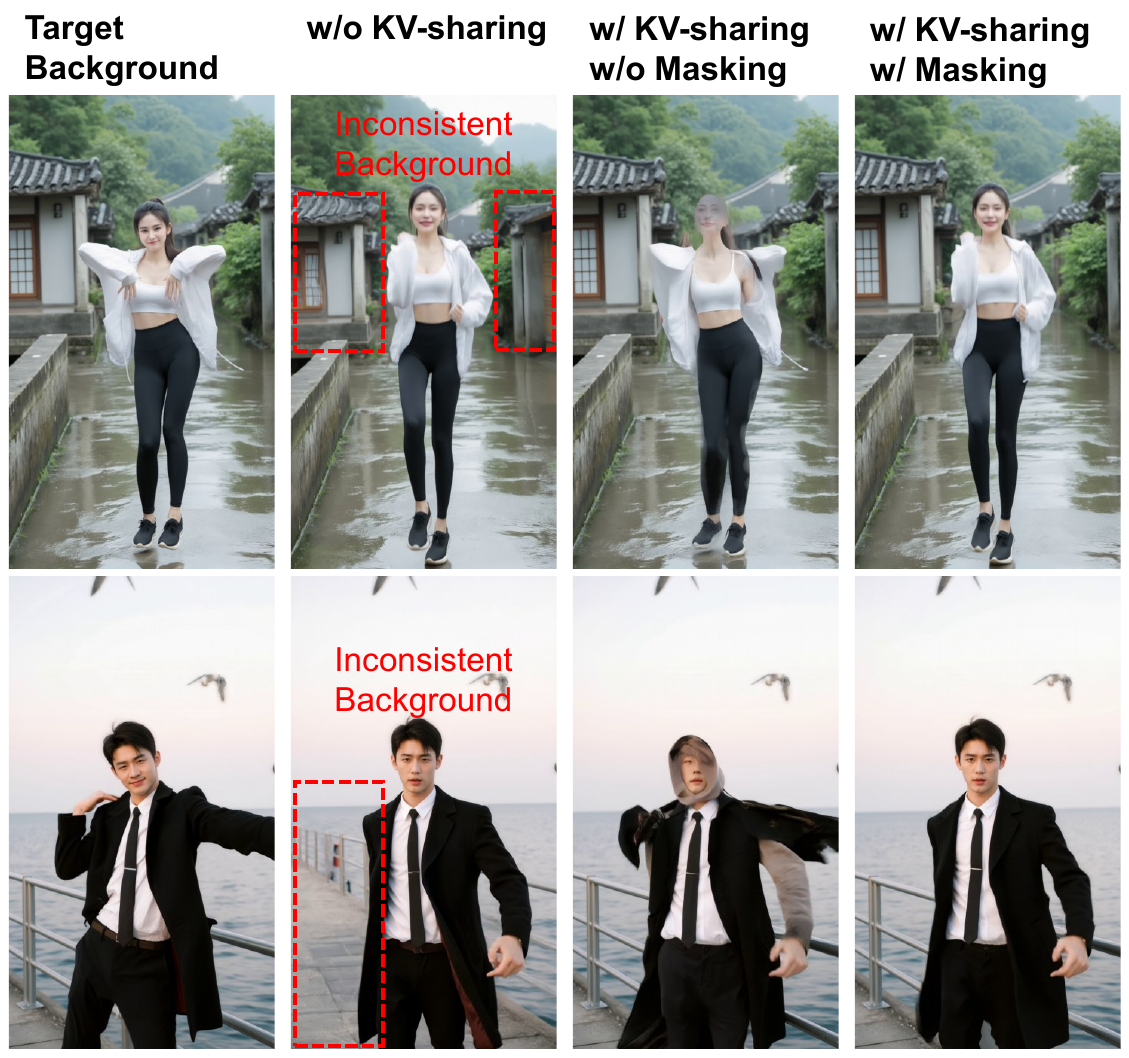}
    \caption{Illustrative ablation on background KV-sharing. Removing KV-sharing leads to background inconsistencies as highlighted in red boxes, while removing masking produces blending artifacts due to conflicting pose signals.}
    \label{fig:illustrative_ablation_kvshare}
\end{figure}

\begin{table}[tb]
    \centering
    \small
    \setlength{\tabcolsep}{3pt}
    \caption{Quantitative ablation on background KV-sharing. KV-sharing enhances temporal consistency, while the masking mechanism restores visual quality and identity preservation, yielding the best overall performance.}
    \begin{tabular}{cc|ccc}
        \toprule
        KV-sharing & Mask & \makecell{Visual \\ Quality} $\uparrow$ & \makecell{Identity \\ Preservation} $\uparrow$ & \makecell{Temporal \\ Consistency} $\uparrow$ \\
        \midrule
        $\times$   & -          & 87.49 & 94.75 & 94.30 \\
        \checkmark & $\times$   & 75.09 & 78.10 & 95.83 \\
        \checkmark & \checkmark & 87.47 & 94.99 & 95.22 \\
        \bottomrule
    \end{tabular}
    \label{tab:quantitative_ablation_kvshare}
\end{table}

\noindent \textbf{Background KV-sharing.}
Background KV-sharing is a pivotal component of our method that facilitates smooth transitions across video segments for coherent long video generation. 
We ablate this component and its attention-based masking mechanism in Fig.~\ref{fig:illustrative_ablation_kvshare}.
As can be seen, removing background KV-sharing results in slight but perceptible background shifts as the woman dances or the man mimics walking. 
While such variations are acceptable within a short video segment, they pose abnormal motions when we stitch multiple segments into a long video. 
Moreover, when the human-background mask derived from cross-attention is removed, the background remains consistent with the target scene; however, the human exhibits noticeable artifacts as a single frame blends two incompatible poses of the same person.
Quantitative results in \cref{tab:quantitative_ablation_kvshare} further demonstrate the importance of background KV-sharing.
Removing KV-sharing leads to a noticeable decrease in temporal consistency (from 95.22 to 94.30).
Although enabling KV-sharing alone improves temporal consistency, it leads to a substantial drop in visual quality and identity preservation, as the human fails to follow the intended pose due to conflicting motion cues propagated from another segment.
These results collectively support the effectiveness of our proposed background KV-sharing in preserving background consistency while faithfully guiding the human's movement according to the current pose.
Since background KV-sharing is performed at initial denoising steps and in deep layers, the impact of attention-based mask accuracy is marginal; we provide a discussion of this in the Appendix.

\begin{figure}[tb]
    \centering
    \includegraphics[width=1.0\linewidth]{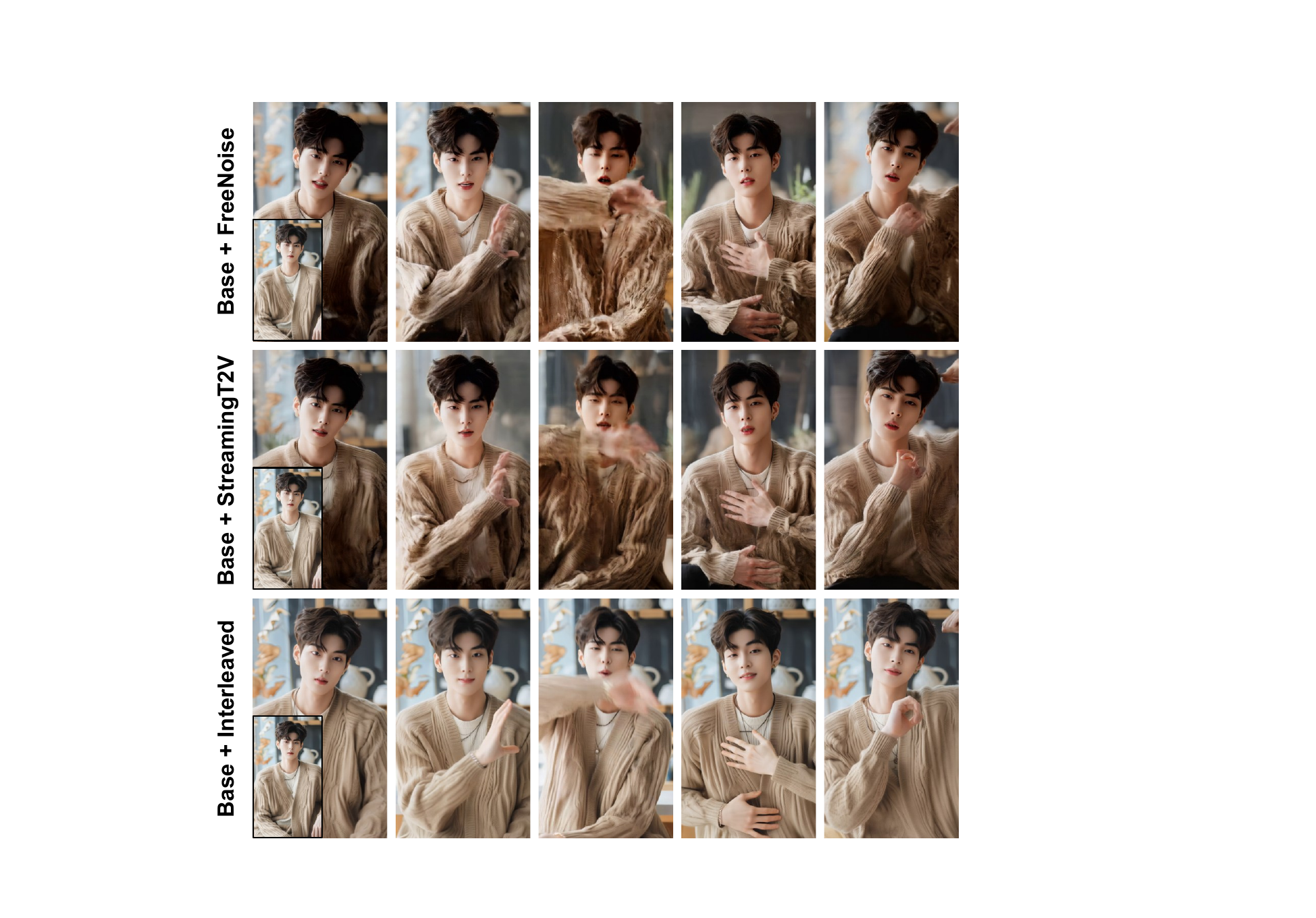}
    \caption{Illustrative comparison results of our interleaved generation framework against alternatives. The reference image is presented at the bottom-left corner. Our method demonstrates strong background consistency across video frames.}
    \label{fig:qualitative_comparison_interleave}
\end{figure}

\begin{table}[tb]
    \centering
    \small
    \setlength{\tabcolsep}{3pt}
    \caption{Quantitative comparison results of our interleaved generation framework for long video generation against alternatives in terms of temporal consistency (frame-wise DINO feature similarity).}
    \begin{tabular}{l|cccc}
        \toprule
        & 10s & 30s & 60s & 90s \\
        \midrule
        Base + FreeNoise~\cite{qiu2023freenoise}            & 0.948	& 0.942 & 0.942 & 0.936 \\
        Base + StreamingT2V~\cite{henschel2025streamingt2v} & 0.948	& 0.942 & 0.941 & 0.934 \\
        Base + Interleaved                                  & 0.951 & 0.952 & 0.952 & 0.951 \\
        \bottomrule
    \end{tabular}
    \label{tab:quantitative_comparison_interleave}
\end{table}

\noindent \textbf{Segment-interleaved generation.}
We compare our segment-interleaved generation framework for long video generation with existing training-free approaches applied to the same base model, including techniques from FreeNoise~\cite{qiu2023freenoise} and StreamingT2V~\cite{henschel2025streamingt2v}, on an additional set of 1-2 minute human dance videos.
As shown in \cref{fig:qualitative_comparison_interleave}, both alternatives suffer from accumulative errors and therefore fail to maintain background consistency.
We further confirm these observations by measuring temporal consistency (frame-wise DINO~\cite{caron2021emerging} feature similarity) in \cref{tab:quantitative_comparison_interleave}.
It is clear that both alternatives exhibit a steady decline in temporal consistency as video length increases from 10s to 90s.
In contrast, our method maintains almost constant scores (0.951-0.952), indicating that background fidelity is well preserved across frames.
These comparisons further demonstrate the effectiveness of our segment-interleaved generation framework in producing temporally coherent long videos.

\begin{table}[tb]
    \centering
    \small
    \setlength{\tabcolsep}{2pt}
    \caption{Ablation on LoRA rank. The values in subscript indicate the relative increase or decrease compared to the adjacent lower-rank setting. Rank 16 provides a favorable trade-off between video quality and computational cost.}
    \begin{tabular}{l|cccc}
        \toprule
        LoRA Rank & 1 & 4 & 16 & 64 \\
        \midrule
        SSIM${\uparrow}$  & 0.60 & 0.61$_{(+0.01)}$ & 0.65$_{(+0.04)}$ & 0.67$_{(+0.02)}$ \\
        FVD${\downarrow}$ & 465  & 443$_{(-22)}$    & 364$_{(-79)}$   & 322$_{(-42)}$ \\
        \bottomrule
    \end{tabular}
    \label{tab:lora_rank}
\end{table}

\noindent \textbf{LoRA rank.}
We conduct ablation on different choices of LoRA rank (1, 4, 16, and 64) using a subset of data as the training and evaluation datasets in our early experiments.
As shown in \cref{tab:lora_rank}, increasing the rank generally improves both image-level (SSIM) and video-level (FVD) metrics.
It is worth noting that rank 16 delivers the most significant gains.
Considering the increased computational cost and memory consumption of higher ranks, we employ a rank of 16 for training on the full dataset, achieving a favorable trade-off between efficiency and video generation quality.
\section{Conclusion}
In this work, we present PoseGen, an effective and efficient framework for generating high-fidelity, pose-controllable human videos of extended duration from a reference image.
By integrating in-context LoRA finetuning with a segment-interleaved generation strategy that leverages background KV-sharing, PoseGen mitigates identity drift and improves temporal consistency.
Extensive experiments show that PoseGen outperforms existing methods in identity preservation, motion control, and temporal coherence, while requiring substantially less training data.
We hope this work provides useful insights for controllable long video generation and inspires future research in this direction.

{
    \small
    \bibliographystyle{ieeenat_fullname}
    \bibliography{main}

\begin{thebibliography}{54}
\providecommand{\natexlab}[1]{#1}
\providecommand{\url}[1]{\texttt{#1}}
\expandafter\ifx\csname urlstyle\endcsname\relax
  \providecommand{\doi}[1]{doi: #1}\else
  \providecommand{\doi}{doi: \begingroup \urlstyle{rm}\Url}\fi

\bibitem[Balaji et~al.(2019)Balaji, Min, Bai, Chellappa, and Graf]{balaji2019conditional}
Yogesh Balaji, Martin~Renqiang Min, Bing Bai, Rama Chellappa, and Hans~Peter Graf.
\newblock Conditional gan with discriminative filter generation for text-to-video synthesis.
\newblock In \emph{IJCAI}, page~2, 2019.

\bibitem[Cai et~al.(2025)Cai, Cun, Li, Liu, Zhang, Zhang, Shan, and Yue]{cai2025ditctrl}
Minghong Cai, Xiaodong Cun, Xiaoyu Li, Wenze Liu, Zhaoyang Zhang, Yong Zhang, Ying Shan, and Xiangyu Yue.
\newblock Ditctrl: Exploring attention control in multi-modal diffusion transformer for tuning-free multi-prompt longer video generation.
\newblock In \emph{Proceedings of the Computer Vision and Pattern Recognition Conference}, pages 7763--7772, 2025.

\bibitem[Cao et~al.(2023)Cao, Wang, Qi, Shan, Qie, and Zheng]{cao2023masactrl}
Mingdeng Cao, Xintao Wang, Zhongang Qi, Ying Shan, Xiaohu Qie, and Yinqiang Zheng.
\newblock Masactrl: Tuning-free mutual self-attention control for consistent image synthesis and editing.
\newblock In \emph{Proceedings of the IEEE/CVF international conference on computer vision}, pages 22560--22570, 2023.

\bibitem[{Cao} et~al.(2019){Cao}, {Hidalgo Martinez}, {Simon}, {Wei}, and {Sheikh}]{cao2019openpose}
Z. {Cao}, G. {Hidalgo Martinez}, T. {Simon}, S. {Wei}, and Y.~A. {Sheikh}.
\newblock Openpose: Realtime multi-person 2d pose estimation using part affinity fields.
\newblock \emph{IEEE Transactions on Pattern Analysis and Machine Intelligence}, 2019.

\bibitem[Caron et~al.(2021)Caron, Touvron, Misra, J{\'e}gou, Mairal, Bojanowski, and Joulin]{caron2021emerging}
Mathilde Caron, Hugo Touvron, Ishan Misra, Herv{\'e} J{\'e}gou, Julien Mairal, Piotr Bojanowski, and Armand Joulin.
\newblock Emerging properties in self-supervised vision transformers.
\newblock In \emph{Proceedings of the IEEE/CVF international conference on computer vision}, pages 9650--9660, 2021.

\bibitem[Chang et~al.(2025)Chang, Xu, Xie, Gao, Kuang, Cai, Zhang, Song, Wang, Shi, et~al.]{chang2025x}
Di Chang, Hongyi Xu, You Xie, Yipeng Gao, Zhengfei Kuang, Shengqu Cai, Chenxu Zhang, Guoxian Song, Chao Wang, Yichun Shi, et~al.
\newblock X-dyna: Expressive dynamic human image animation.
\newblock In \emph{Proceedings of the IEEE/CVF Conference on Computer Vision and Pattern Recognition}, pages 5499--5509, 2025.

\bibitem[Chen et~al.(2025{\natexlab{a}})Chen, Lin, Yang, Lin, Zhu, Fan, Zhang, Chen, Chen, Ma, et~al.]{chen2025skyreels}
Guibin Chen, Dixuan Lin, Jiangping Yang, Chunze Lin, Junchen Zhu, Mingyuan Fan, Hao Zhang, Sheng Chen, Zheng Chen, Chengcheng Ma, et~al.
\newblock Skyreels-v2: Infinite-length film generative model.
\newblock \emph{arXiv preprint arXiv:2504.13074}, 2025{\natexlab{a}}.

\bibitem[Chen et~al.(2019)Chen, Wang, Pang, Cao, Xiong, Li, Sun, Feng, Liu, Xu, et~al.]{chen2019mmdetection}
Kai Chen, Jiaqi Wang, Jiangmiao Pang, Yuhang Cao, Yu Xiong, Xiaoxiao Li, Shuyang Sun, Wansen Feng, Ziwei Liu, Jiarui Xu, et~al.
\newblock Mmdetection: Open mmlab detection toolbox and benchmark.
\newblock \emph{arXiv preprint arXiv:1906.07155}, 2019.

\bibitem[Chen et~al.(2025{\natexlab{b}})Chen, Liang, Zhou, Huang, Ma, Tang, Lin, Zhou, and Lu]{hu2025HunyuanVideo-Avatar}
Yi Chen, Sen Liang, Zixiang Zhou, Ziyao Huang, Yifeng Ma, Junshu Tang, Qin Lin, Yuan Zhou, and Qinglin Lu.
\newblock Hunyuanvideo-avatar: High-fidelity audio-driven human animation for multiple characters, 2025{\natexlab{b}}.

\bibitem[Deng et~al.(2019)Deng, Guo, Xue, and Zafeiriou]{deng2019arcface}
Jiankang Deng, Jia Guo, Niannan Xue, and Stefanos Zafeiriou.
\newblock Arcface: Additive angular margin loss for deep face recognition.
\newblock In \emph{Proceedings of the IEEE/CVF conference on computer vision and pattern recognition}, pages 4690--4699, 2019.

\bibitem[Gan et~al.(2025)Gan, Ren, Zhang, Ye, Xie, Yin, Yuan, Peng, and Zhu]{gan2025humandit}
Qijun Gan, Yi Ren, Chen Zhang, Zhenhui Ye, Pan Xie, Xiang Yin, Zehuan Yuan, Bingyue Peng, and Jianke Zhu.
\newblock Humandit: Pose-guided diffusion transformer for long-form human motion video generation.
\newblock \emph{arXiv preprint arXiv:2502.04847}, 2025.

\bibitem[Guo et~al.(2024)Guo, Yang, Rao, Liang, Wang, Qiao, Agrawala, Lin, and Dai]{guo2023animatediff}
Yuwei Guo, Ceyuan Yang, Anyi Rao, Zhengyang Liang, Yaohui Wang, Yu Qiao, Maneesh Agrawala, Dahua Lin, and Bo Dai.
\newblock Animatediff: Animate your personalized text-to-image diffusion models without specific tuning.
\newblock \emph{International Conference on Learning Representations}, 2024.

\bibitem[Henschel et~al.(2025)Henschel, Khachatryan, Poghosyan, Hayrapetyan, Tadevosyan, Wang, Navasardyan, and Shi]{henschel2025streamingt2v}
Roberto Henschel, Levon Khachatryan, Hayk Poghosyan, Daniil Hayrapetyan, Vahram Tadevosyan, Zhangyang Wang, Shant Navasardyan, and Humphrey Shi.
\newblock Streamingt2v: Consistent, dynamic, and extendable long video generation from text.
\newblock In \emph{Proceedings of the Computer Vision and Pattern Recognition Conference}, pages 2568--2577, 2025.

\bibitem[Heusel et~al.(2017)Heusel, Ramsauer, Unterthiner, Nessler, and Hochreiter]{heusel2017gans}
Martin Heusel, Hubert Ramsauer, Thomas Unterthiner, Bernhard Nessler, and Sepp Hochreiter.
\newblock Gans trained by a two time-scale update rule converge to a local nash equilibrium.
\newblock \emph{Advances in neural information processing systems}, 30, 2017.

\bibitem[Ho et~al.(2020)Ho, Jain, and Abbeel]{ho2020denoising}
Jonathan Ho, Ajay Jain, and Pieter Abbeel.
\newblock Denoising diffusion probabilistic models.
\newblock \emph{Advances in neural information processing systems}, 33:\penalty0 6840--6851, 2020.

\bibitem[Hore and Ziou(2010)]{hore2010image}
Alain Hore and Djemel Ziou.
\newblock Image quality metrics: Psnr vs. ssim.
\newblock In \emph{2010 20th international conference on pattern recognition}, pages 2366--2369. IEEE, 2010.

\bibitem[Hu et~al.(2022)Hu, Shen, Wallis, Allen-Zhu, Li, Wang, Wang, Chen, et~al.]{hu2022lora}
Edward~J Hu, Yelong Shen, Phillip Wallis, Zeyuan Allen-Zhu, Yuanzhi Li, Shean Wang, Lu Wang, Weizhu Chen, et~al.
\newblock Lora: Low-rank adaptation of large language models.
\newblock \emph{ICLR}, 1\penalty0 (2):\penalty0 3, 2022.

\bibitem[Hu(2024)]{hu2024animate}
Li Hu.
\newblock Animate anyone: Consistent and controllable image-to-video synthesis for character animation.
\newblock In \emph{Proceedings of the IEEE/CVF Conference on Computer Vision and Pattern Recognition}, pages 8153--8163, 2024.

\bibitem[Hu et~al.(2025)Hu, Yu, Zhou, Liang, Zhou, Lin, and Lu]{hu2025hunyuancustom}
Teng Hu, Zhentao Yu, Zhengguang Zhou, Sen Liang, Yuan Zhou, Qin Lin, and Qinglin Lu.
\newblock Hunyuancustom: A multimodal-driven architecture for customized video generation.
\newblock \emph{arXiv preprint arXiv:2505.04512}, 2025.

\bibitem[Jafarian and Park(2021)]{jafarian2021learning}
Yasamin Jafarian and Hyun~Soo Park.
\newblock Learning high fidelity depths of dressed humans by watching social media dance videos.
\newblock In \emph{Proceedings of the IEEE/CVF Conference on Computer Vision and Pattern Recognition}, pages 12753--12762, 2021.

\bibitem[Jiang et~al.(2025)Jiang, Han, Mao, Zhang, Pan, and Liu]{jiang2025vace}
Zeyinzi Jiang, Zhen Han, Chaojie Mao, Jingfeng Zhang, Yulin Pan, and Yu Liu.
\newblock Vace: All-in-one video creation and editing.
\newblock \emph{arXiv preprint arXiv:2503.07598}, 2025.

\bibitem[Jocher et~al.(2022)Jocher, Chaurasia, Stoken, Borovec, Kwon, Michael, Fang, Yifu, Wong, Montes, et~al.]{ultralytics}
Glenn Jocher, Ayush Chaurasia, Alex Stoken, Jirka Borovec, Yonghye Kwon, Kalen Michael, Jiacong Fang, Zeng Yifu, Colin Wong, Diego Montes, et~al.
\newblock ultralytics/yolov5: v7.0 - yolov5 sota realtime instance segmentation.
\newblock \emph{Zenodo}, 2022.

\bibitem[Khirodkar et~al.(2024)Khirodkar, Bagautdinov, Martinez, Zhaoen, James, Selednik, Anderson, and Saito]{khirodkar2024sapiens}
Rawal Khirodkar, Timur Bagautdinov, Julieta Martinez, Su Zhaoen, Austin James, Peter Selednik, Stuart Anderson, and Shunsuke Saito.
\newblock Sapiens: Foundation for human vision models.
\newblock In \emph{European Conference on Computer Vision}, pages 206--228. Springer, 2024.

\bibitem[Kim et~al.(2024)Kim, Kang, Choi, and Han]{kim2024fifo}
Jihwan Kim, Junoh Kang, Jinyoung Choi, and Bohyung Han.
\newblock Fifo-diffusion: Generating infinite videos from text without training.
\newblock \emph{Advances in Neural Information Processing Systems}, 37:\penalty0 89834--89868, 2024.

\bibitem[Li et~al.(2024)Li, Li, Yang, Cao, Zhu, Cheng, and Chen]{li2024dispose}
Hongxiang Li, Yaowei Li, Yuhang Yang, Junjie Cao, Zhihong Zhu, Xuxin Cheng, and Long Chen.
\newblock Dispose: Disentangling pose guidance for controllable human image animation.
\newblock \emph{arXiv preprint arXiv:2412.09349}, 2024.

\bibitem[Lin et~al.(2025)Lin, Jiang, Yang, Zheng, and Liang]{lin2025omnihuman}
Gaojie Lin, Jianwen Jiang, Jiaqi Yang, Zerong Zheng, and Chao Liang.
\newblock Omnihuman-1: Rethinking the scaling-up of one-stage conditioned human animation models.
\newblock \emph{arXiv preprint arXiv:2502.01061}, 2025.

\bibitem[Luo et~al.(2025)Luo, Rong, Wang, Zhang, Hu, and Zhu]{luo2025dreamactor}
Yuxuan Luo, Zhengkun Rong, Lizhen Wang, Longhao Zhang, Tianshu Hu, and Yongming Zhu.
\newblock Dreamactor-m1: Holistic, expressive and robust human image animation with hybrid guidance.
\newblock \emph{arXiv preprint arXiv:2504.01724}, 2025.

\bibitem[Peebles and Xie(2022)]{peebles2022DiT}
William Peebles and Saining Xie.
\newblock Scalable diffusion models with transformers.
\newblock \emph{arXiv preprint arXiv:2212.09748}, 2022.

\bibitem[Peebles and Xie(2023)]{peebles2023scalable}
William Peebles and Saining Xie.
\newblock Scalable diffusion models with transformers.
\newblock In \emph{Proceedings of the IEEE/CVF international conference on computer vision}, pages 4195--4205, 2023.

\bibitem[Qiu et~al.(2023)Qiu, Xia, Zhang, He, Wang, Shan, and Liu]{qiu2023freenoise}
Haonan Qiu, Menghan Xia, Yong Zhang, Yingqing He, Xintao Wang, Ying Shan, and Ziwei Liu.
\newblock Freenoise: Tuning-free longer video diffusion via noise rescheduling.
\newblock \emph{arXiv preprint arXiv:2310.15169}, 2023.

\bibitem[Radford et~al.(2021)Radford, Kim, Hallacy, Ramesh, Goh, Agarwal, Sastry, Askell, Mishkin, Clark, et~al.]{radford2021learning}
Alec Radford, Jong~Wook Kim, Chris Hallacy, Aditya Ramesh, Gabriel Goh, Sandhini Agarwal, Girish Sastry, Amanda Askell, Pamela Mishkin, Jack Clark, et~al.
\newblock Learning transferable visual models from natural language supervision.
\newblock In \emph{International conference on machine learning}, pages 8748--8763. PmLR, 2021.

\bibitem[Rombach et~al.(2022)Rombach, Blattmann, Lorenz, Esser, and Ommer]{rombach2022high}
Robin Rombach, Andreas Blattmann, Dominik Lorenz, Patrick Esser, and Bj{\"o}rn Ommer.
\newblock High-resolution image synthesis with latent diffusion models.
\newblock In \emph{Proceedings of the IEEE/CVF Conference on Computer Vision and Pattern Recognition (CVPR)}, pages 10684--10695, 2022.

\bibitem[Ruiz et~al.(2023)Ruiz, Li, Jampani, Pritch, Rubinstein, and Aberman]{ruiz2023dreambooth}
Nataniel Ruiz, Yuanzhen Li, Varun Jampani, Yael Pritch, Michael Rubinstein, and Kfir Aberman.
\newblock Dreambooth: Fine tuning text-to-image diffusion models for subject-driven generation.
\newblock In \emph{Proceedings of the IEEE/CVF Conference on Computer Vision and Pattern Recognition}, 2023.

\bibitem[Sohl-Dickstein et~al.(2015)Sohl-Dickstein, Weiss, Maheswaranathan, and Ganguli]{sohl2015deep}
Jascha Sohl-Dickstein, Eric Weiss, Niru Maheswaranathan, and Surya Ganguli.
\newblock Deep unsupervised learning using nonequilibrium thermodynamics.
\newblock In \emph{International conference on machine learning}, pages 2256--2265. pmlr, 2015.

\bibitem[Su et~al.(2024)Su, Ahmed, Lu, Pan, Bo, and Liu]{su2024roformer}
Jianlin Su, Murtadha Ahmed, Yu Lu, Shengfeng Pan, Wen Bo, and Yunfeng Liu.
\newblock Roformer: Enhanced transformer with rotary position embedding.
\newblock \emph{Neurocomputing}, 568:\penalty0 127063, 2024.

\bibitem[Tan et~al.(2024)Tan, Liu, Yang, Xue, and Wang]{tan2024ominicontrol}
Zhenxiong Tan, Songhua Liu, Xingyi Yang, Qiaochu Xue, and Xinchao Wang.
\newblock Ominicontrol: Minimal and universal control for diffusion transformer.
\newblock \emph{arXiv preprint arXiv:2411.15098}, 2024.

\bibitem[Teng et~al.(2025)Teng, Jia, Sun, Li, Li, Tang, Han, Zhang, Zhang, Luo, et~al.]{teng2025magi}
Hansi Teng, Hongyu Jia, Lei Sun, Lingzhi Li, Maolin Li, Mingqiu Tang, Shuai Han, Tianning Zhang, WQ Zhang, Weifeng Luo, et~al.
\newblock Magi-1: Autoregressive video generation at scale.
\newblock \emph{arXiv preprint arXiv:2505.13211}, 2025.

\bibitem[Tu et~al.(2025)Tu, Xing, Han, Cheng, Dai, Luo, and Wu]{tu2025stableanimator}
Shuyuan Tu, Zhen Xing, Xintong Han, Zhi-Qi Cheng, Qi Dai, Chong Luo, and Zuxuan Wu.
\newblock Stableanimator: High-quality identity-preserving human image animation.
\newblock In \emph{Proceedings of the Computer Vision and Pattern Recognition Conference}, pages 21096--21106, 2025.

\bibitem[Unterthiner et~al.(2018)Unterthiner, Van~Steenkiste, Kurach, Marinier, Michalski, and Gelly]{unterthiner2018towards}
Thomas Unterthiner, Sjoerd Van~Steenkiste, Karol Kurach, Raphael Marinier, Marcin Michalski, and Sylvain Gelly.
\newblock Towards accurate generative models of video: A new metric \& challenges.
\newblock \emph{arXiv preprint arXiv:1812.01717}, 2018.

\bibitem[Vaswani et~al.(2017)Vaswani, Shazeer, Parmar, Uszkoreit, Jones, Gomez, Kaiser, and Polosukhin]{vaswani2017attention}
Ashish Vaswani, Noam Shazeer, Niki Parmar, Jakob Uszkoreit, Llion Jones, Aidan~N Gomez, {\L}ukasz Kaiser, and Illia Polosukhin.
\newblock Attention is all you need.
\newblock \emph{Advances in neural information processing systems}, 30, 2017.

\bibitem[Wan et~al.(2025)Wan, Wang, Ai, Wen, Mao, Xie, Chen, Yu, Zhao, Yang, et~al.]{wan2025wan}
Team Wan, Ang Wang, Baole Ai, Bin Wen, Chaojie Mao, Chen-Wei Xie, Di Chen, Feiwu Yu, Haiming Zhao, Jianxiao Yang, et~al.
\newblock Wan: Open and advanced large-scale video generative models.
\newblock \emph{arXiv preprint arXiv:2503.20314}, 2025.

\bibitem[Wang et~al.(2024)Wang, Zhang, Gao, Wang, Zhou, Zhang, Yan, and Sang]{wang2024unianimate}
Xiang Wang, Shiwei Zhang, Changxin Gao, Jiayu Wang, Xiaoqiang Zhou, Yingya Zhang, Luxin Yan, and Nong Sang.
\newblock Unianimate: Taming unified video diffusion models for consistent human image animation.
\newblock \emph{arXiv preprint arXiv:2406.01188}, 2024.

\bibitem[Wang et~al.(2025)Wang, Zhang, Tang, Zhang, Gao, Wang, and Sang]{wang2025unianimate}
Xiang Wang, Shiwei Zhang, Longxiang Tang, Yingya Zhang, Changxin Gao, Yuehuan Wang, and Nong Sang.
\newblock Unianimate-dit: Human image animation with large-scale video diffusion transformer.
\newblock \emph{arXiv preprint arXiv:2504.11289}, 2025.

\bibitem[Wang et~al.(2004)Wang, Bovik, Sheikh, and Simoncelli]{wang2004image}
Zhou Wang, Alan~C Bovik, Hamid~R Sheikh, and Eero~P Simoncelli.
\newblock Image quality assessment: from error visibility to structural similarity.
\newblock \emph{IEEE transactions on image processing}, 13\penalty0 (4):\penalty0 600--612, 2004.

\bibitem[Wu et~al.(2023)Wu, Zhang, Zhang, Chen, Liao, Li, Gao, Wang, Zhang, Sun, et~al.]{wu2023q}
Haoning Wu, Zicheng Zhang, Weixia Zhang, Chaofeng Chen, Liang Liao, Chunyi Li, Yixuan Gao, Annan Wang, Erli Zhang, Wenxiu Sun, et~al.
\newblock Q-align: Teaching lmms for visual scoring via discrete text-defined levels.
\newblock \emph{arXiv preprint arXiv:2312.17090}, 2023.

\bibitem[Yang et~al.(2024{\natexlab{a}})Yang, Yang, Hui, Zheng, Yu, Zhou, Li, Li, Liu, Huang, et~al.]{qwen2}
An Yang, Baosong Yang, Binyuan Hui, Bo Zheng, Bowen Yu, Chang Zhou, Chengpeng Li, Chengyuan Li, Dayiheng Liu, Fei Huang, et~al.
\newblock Qwen2 technical report.
\newblock \emph{arXiv preprint arXiv:2407.10671}, 2024{\natexlab{a}}.

\bibitem[Yang et~al.(2023)Yang, Zeng, Yuan, and Li]{yang2023effective}
Zhendong Yang, Ailing Zeng, Chun Yuan, and Yu Li.
\newblock Effective whole-body pose estimation with two-stages distillation.
\newblock In \emph{Proceedings of the IEEE/CVF International Conference on Computer Vision}, pages 4210--4220, 2023.

\bibitem[Yang et~al.(2024{\natexlab{b}})Yang, Teng, Zheng, Ding, Huang, Xu, Yang, Hong, Zhang, Feng, et~al.]{yang2024cogvideox}
Zhuoyi Yang, Jiayan Teng, Wendi Zheng, Ming Ding, Shiyu Huang, Jiazheng Xu, Yuanming Yang, Wenyi Hong, Xiaohan Zhang, Guanyu Feng, et~al.
\newblock Cogvideox: Text-to-video diffusion models with an expert transformer.
\newblock \emph{arXiv preprint arXiv:2408.06072}, 2024{\natexlab{b}}.

\bibitem[Yin et~al.(2025)Yin, Zhang, Zhang, Freeman, Durand, Shechtman, and Huang]{yin2025slow}
Tianwei Yin, Qiang Zhang, Richard Zhang, William~T Freeman, Fredo Durand, Eli Shechtman, and Xun Huang.
\newblock From slow bidirectional to fast autoregressive video diffusion models.
\newblock In \emph{Proceedings of the Computer Vision and Pattern Recognition Conference}, pages 22963--22974, 2025.

\bibitem[Zhang et~al.(2023)Zhang, Rao, and Agrawala]{zhang2023adding}
Lvmin Zhang, Anyi Rao, and Maneesh Agrawala.
\newblock Adding conditional control to text-to-image diffusion models.
\newblock In \emph{Proceedings of the IEEE/CVF international conference on computer vision}, pages 3836--3847, 2023.

\bibitem[Zhang et~al.(2018)Zhang, Isola, Efros, Shechtman, and Wang]{zhang2018unreasonable}
Richard Zhang, Phillip Isola, Alexei~A Efros, Eli Shechtman, and Oliver Wang.
\newblock The unreasonable effectiveness of deep features as a perceptual metric.
\newblock In \emph{Proceedings of the IEEE conference on computer vision and pattern recognition}, pages 586--595, 2018.

\bibitem[Zhang et~al.(2025)Zhang, Gu, Wang, Wang, Cheng, Zhu, and Zou]{zhang2025mimicmotion}
Yuang Zhang, Jiaxi Gu, Li-Wen Wang, Han Wang, Junqi Cheng, Yuefeng Zhu, and Fangyuan Zou.
\newblock Mimicmotion: High-quality human motion video generation with confidence-aware pose guidance.
\newblock In \emph{International Conference on Machine Learning}, 2025.

\bibitem[Zhou et~al.(2024)Zhou, Wang, Chen, Bai, Li, Zhang, Xu, Yang, and Wang]{zhou2024realisdance}
Jingkai Zhou, Benzhi Wang, Weihua Chen, Jingqi Bai, Dongyang Li, Aixi Zhang, Hao Xu, Mingyang Yang, and Fan Wang.
\newblock Realisdance: Equip controllable character animation with realistic hands.
\newblock \emph{arXiv preprint arXiv:2409.06202}, 2024.

\bibitem[Zhou et~al.(2025)Zhou, Wu, Li, Wei, Fan, Chen, Jiang, and Wang]{zhou2025realisdance}
Jingkai Zhou, Yifan Wu, Shikai Li, Min Wei, Chao Fan, Weihua Chen, Wei Jiang, and Fan Wang.
\newblock Realisdance-dit: Simple yet strong baseline towards controllable character animation in the wild.
\newblock \emph{arXiv preprint arXiv:2504.14977}, 2025.

\end{thebibliography}
}

\clearpage

\setcounter{section}{0}
\setcounter{figure}{0}
\setcounter{table}{0}
\setcounter{equation}{0}

\renewcommand{\thesection}{S\arabic{section}}
\renewcommand{\thefigure}{S\arabic{figure}}
\renewcommand{\thetable}{S\arabic{table}}
\renewcommand{\theequation}{S\arabic{equation}}

\clearpage
\setcounter{page}{1}
\maketitlesupplementary

In this supplementary material, we present more details of our training data, implementation, additional qualitative results, ablation studies, robustness of hand normals, inference efficiency, and limitations.

\begin{figure*}[htbp]
    \centering
    \includegraphics[width=1.0\linewidth]{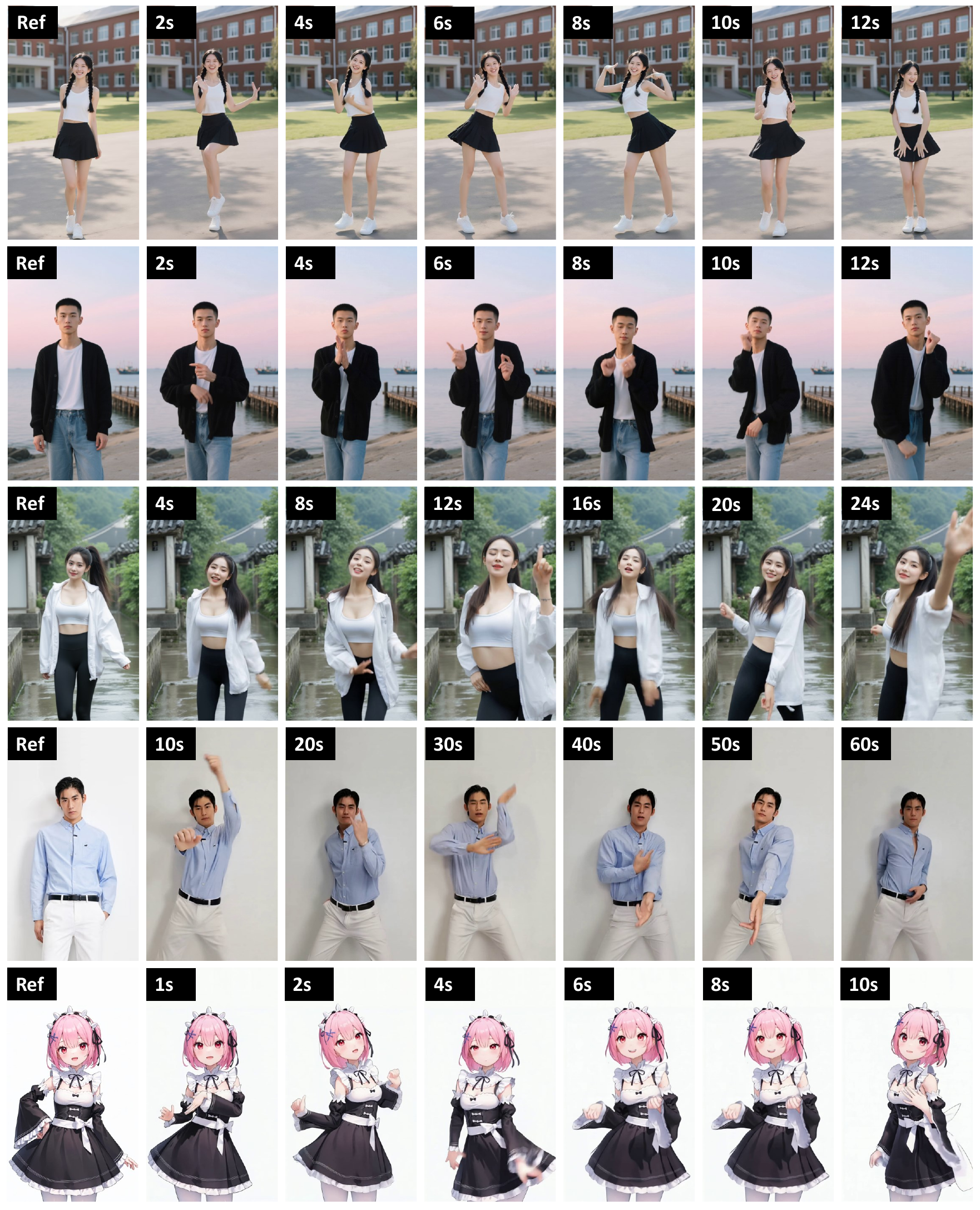}
    \caption{Additional qualitative results of pose-controllable human video generation. The reference images are displayed at the first column. PoseGen is capable of animating cartoon-style images as depicted in the last row.}
    \label{fig:results}
\end{figure*}

\begin{figure*}[htbp]
    \centering
    \includegraphics[width=0.9\linewidth]{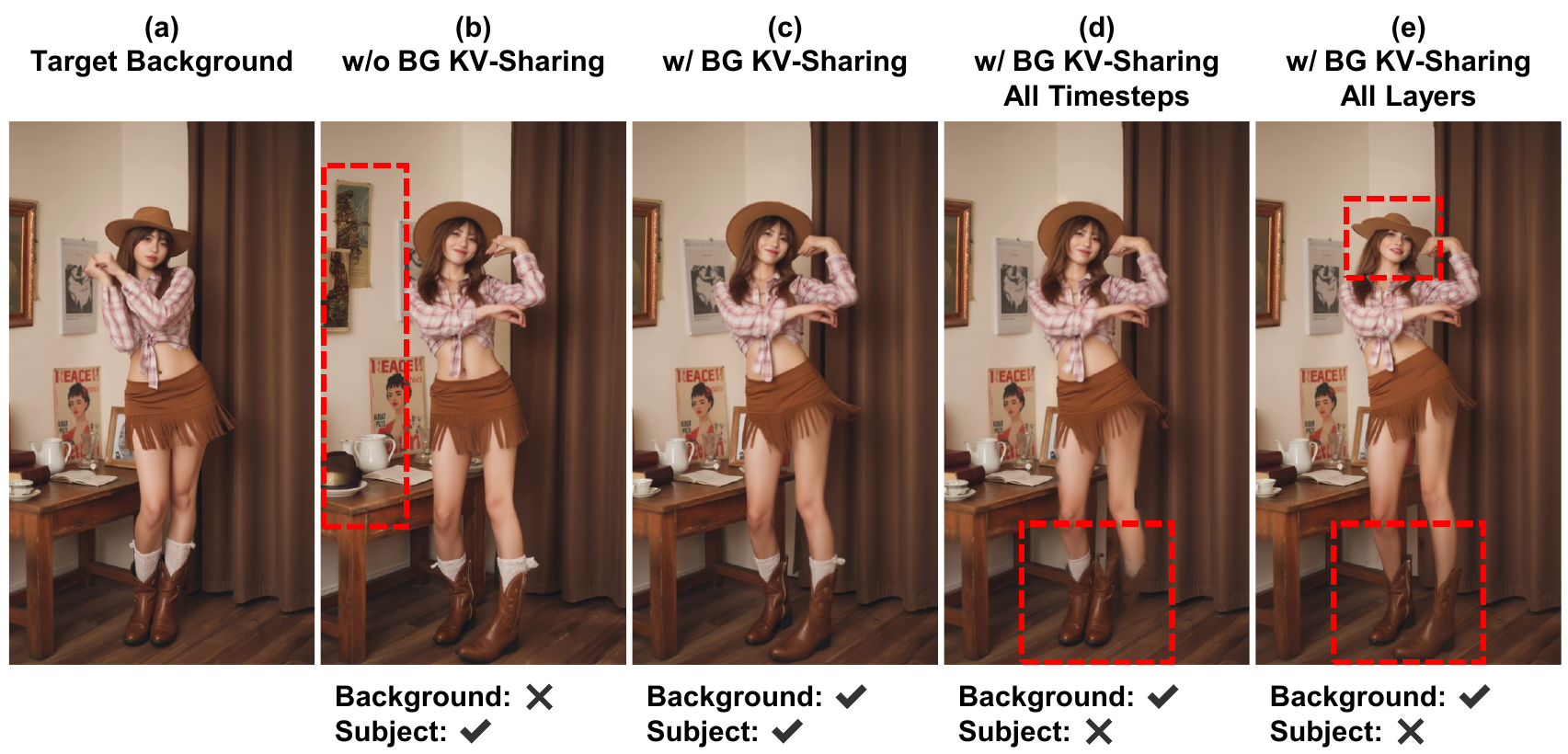}
    \caption{Ablation study on background KV-sharing in terms of denoising timesteps and transformer layers. (a) The source segment with the target background. (b) Background KV-sharing is not applied. (c) Background KV-sharing is applied to the first five timesteps and the last ten layers. (d) Background KV-sharing is applied to all timesteps and the last ten layers. (e) Background KV-sharing is applied to the first five timesteps and all layers.}
    \label{fig:kvsharing}
\end{figure*}

\section{Training Data}

\subsection{Data Construction}
We collect human-centric videos that capture the movement of a single person, such as dancing or live broadcasting, from various online platforms. 
To ensure the quality of these videos for training, we employ a comprehensive filtering process. Initially, the collected videos undergo a manual screening to discard samples containing undesirable elements such as abrupt scene changes, background jitter, or superimposed special effects. Following this, we apply an automated filtering pipeline to only preserve samples that simultaneously meet the following criteria. 
(1) \textbf{Visual Quality.}
We adopt the image and video quality scorers provided by Q-Align~\cite{wu2023q} to assess the visual quality of a collected video. 
A threshold of 0.7 is applied for both scorers to exclude low-quality samples. 
(2) \textbf{Human Presence.}
We uniformly sample a set of frames from a video and extract human bounding boxes using RTMDet~\cite{chen2019mmdetection}. 
We then compute the human area ratio according to the estimated bounding boxes. 
Videos with an average ratio below 0.3 are discarded to eliminate those with insufficient human presence. 
Meanwhile, this moderate threshold effectively retains videos where humans are captured in close-up, medium, or full shots. 
(3) \textbf{Facial Confidence.}
We apply YOLO~\cite{ultralytics} to extract bounding box annotations for the categories ``face'' and ``head'' in each sampled frame. 
Videos with an average confidence score exceeding 0.75 are retained to ensure clear visibility of human face and head regions.
This filtering choice helps avoid significant occlusions from masks, hats, or similar accessories, which can detrimentally affect the performance of identity-preserving human video generation. 
Finally, we obtain a training dataset that comprises 7,005 high-quality video clips for LoRA finetuning. 

\subsection{Data Preprocessing}
We leverage Sapiens~\cite{khirodkar2024sapiens}, a family of human vision models, to predict pose skeleton maps and hand normal maps for training videos. 
Regarding pose estimation, the score threshold and the Intersection-over-Union (IoU) threshold for human bounding box detection are both set to 0.3. 
We render OpenPose~\cite{cao2019openpose}-style body and hand keypoints while discarding facial keypoints, in order to avoid identity leakage from the driving video to the generated subject.
To balance precision and recall, we empirically set the confidence threshold to 0.3 for body keypoints and 0.7 for hand keypoints to ensure accurate pose skeletons.
For hand normal prediction, we first apply the surface normal prediction model of Sapiens to generate normal maps for the entire frame, and then use its body-part segmentation model to localize the desired hand regions. 
Considering that the quality of hand regions in a training video is often affected by motion blurs, a dropout rate of 0.1 is applied to both hand skeletons and surface normals in order to decouple the strong reliance on these conditions and promote model robustness. 
We adopt Qwen2.5-7B-Instruct~\cite{qwen2} as our text encoder to generate text prompts for reference images. 
The structure of a prompt typically begins with the image style, followed by a detailed description of the subject and the background, and ends with the type of the camera shot, which is similar to the prompt distribution of Wan2.1 post-training captions. 

\section{Additional Implementation Details}
Our model is built upon the checkpoint of Wan2.1 Image-to-Video (I2V) 14B model~\cite{wan2025wan}. 
The learnable parameters in the reference image patchifier are initialized by copying those in the noisy video patchifier. 
To prevent color-related artifacts that we observed in early experiments when concatenating hand normal maps with pose skeletons, we devised an alternative integration strategy. Instead of a simple concatenation, we process the hand normal maps through a separate patchifier and then add its output to the pose skeleton features. This convolution-addition scheme effectively fuses geometric information without introducing color spillover. 
We then append a zero-initialized linear projection layer after the 3D convolutional layer in the separated hand patchifier to suppress its influence at the beginning of the training process, inspired by the design of ControlNet~\cite{zhang2023adding}.
During training, the learning rate is linearly warmed up to $1e-4$ over the first 10\% of the total training steps, followed by a cosine annealing scheduler for the remaining steps. 
The LoRA rank and alpha are both set to $16$. 

\section{Additional Qualitative Results}
We present more qualitative results of pose-driven human video generation in Fig.~\ref{fig:results}. These visualizations showcase that our model effectively synthesizes realistic human videos while preserving identity consistency across diverse scenes and complex motion patterns. Notably, PoseGen exhibits impressive generalization capabilities beyond the real-human videos it was trained on. As demonstrated in the bottom row of Fig.~\ref{fig:results}, by first applying a pose retargeting process~\cite{wang2025unianimate} to the driving motion, our model can successfully animate out-of-domain subjects like cartoon characters, faithfully transferring the motion while preserving the distinct artistic style.

\section{Additional Ablation Study}
\label{sec:suppl_ablation}
Following the settings in DiTCtrl~\cite{cai2025ditctrl}, we perform background KV-sharing at early denoising timesteps and at deep transformer layers. 
To further evaluate the effect of denoising timesteps and transformer layers, we conducted an ablation on the background KV-sharing mechanism, and the results are presented in Fig.~\ref{fig:kvsharing}. 
As shown in Fig.~\ref{fig:kvsharing}, background KV-sharing not only preserves the desired background from the source segment when generating the current segment, but also successfully drives the subject according to the pose signals without introducing noticeable artifacts. 
Specifically, applying background KV-sharing across all denoising timesteps causes inadvertent propagation of human pose information from the source to the current segment, resulting in overlapping poses and reduced motion fidelity. 
In contrast, implementing background KV-sharing across all transformer layers allows low-level visual features to leak from the source segment, leading to visual artifacts on the subject.
These observations corroborate the findings in MasaCtrl~\cite{cao2023masactrl}, which indicate that source background is primarily conveyed during initial denoising steps and in deeper network layers.
Our approach retains the first denoising step, a minor deviation from DiTCtrl~\cite{cai2025ditctrl} and MasaCtrl~\cite{cao2023masactrl}, to ensure consistency for the identical background expected across video segments.

\section{Robustness of Hand Normals}

\begin{figure}[tb]
    \centering
    \includegraphics[width=\linewidth]{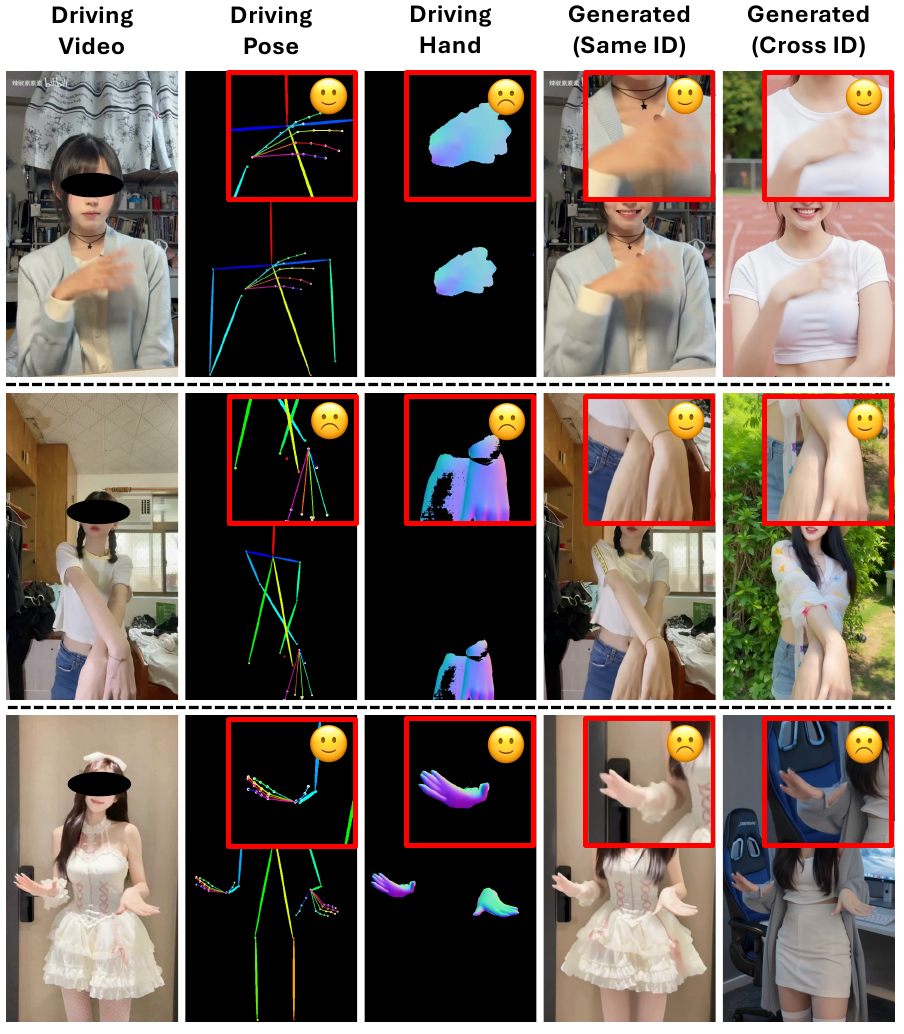}
    \caption{\textbf{Illustrations of hand surface normal robustness.} In the 1st row, despite inaccurate normal caused by motion blur, the model generates plausible hands by referring to pose. In the 2nd row, even when both pose and normal are inaccurate, our method can generate reasonable hands thanks to temporal modeling. The 3rd row presents a failure case where complex clothing interferes with hand appearance despite accurate conditions.}
    \label{fig:hand_robustness}
\end{figure}

As illustrated in \cref{fig:hand_robustness}, the generation quality is robust to inaccurate hand normals, as our framework jointly leverages pose skeletons, hand normals, and the network's inherent temporal modeling capability to generate videos.

\section{Inference Efficiency}

\begin{table}[tb]
    \centering
    \small
    \begin{tabular}{l|ccc}
    \toprule
    & UniAnimate-DiT & PoseGen$^{1}$ & PoseGen$^{2}$ \\
    \midrule
    Time (s/it) & 64.9 / 69.5 & 75.0 / 83.3 & 71.2 / 77.5 \\
    Memory (G)  & 54.1 / 25.3 & 57.0 / 27.1 & 45.3 / 14.7 \\
    \bottomrule
    \end{tabular}
    \caption{\textbf{Inference efficiency}. PoseGen$^{1}$ denotes generating non-overlapping segments with KV-sharing. PoseGen$^{2}$ denotes generating in-between segments without KV-sharing. Each cell value is presented as ``without / with'' memory management. We run the test at $1280 \times 720$ resolution on a A100 GPU. Our method exhibits comparable runtime and manageable memory usage.}
    \label{tab:inference}
\end{table}

We compare inference cost with UniAnimate-DiT~\cite{wang2025unianimate}, which adopts a similar architecture, in \cref{tab:inference}.
For segment-interleaved generation, the runtime overhead mainly comes from 1/4 overlapping between adjacent segments that we designed for better temporal coherence, while the memory overhead stems from caching key-value pairs on GPU for background KV-sharing.

\section{Limitations}
Despite its strong performance, PoseGen also exhibits a few limitations. 
First, the interleaved segment-wise generation approach may introduce subtle drift in fine-grained details (such as the subject's delicate accessories) or in occluded regions of the reference image.
Second, the model's ability to create seamless long videos hinges on the selected source segment having a static background, as any source variations can lead to unnatural artifacts in the segments to be generated.
Third, in terms of controllability, our framework does not support nuanced control of facial expressions, which we leave for future work.


\end{document}